\documentclass[10pt,twocolumn,letterpaper]{article}

\usepackage[pagenumbers]{iccv} 

\definecolor{iccvblue}{rgb}{0.21,0.49,0.74}
\usepackage[pagebackref,breaklinks,colorlinks,allcolors=iccvblue]{hyperref}
\usepackage{amsthm}

\usepackage{amsmath,amsfonts,bm}

\def\eqref#1{equation~\ref{#1}}

\def\1{\bm{1}}

\def\rd{{\textnormal{d}}}

\def\vx{{\bm{x}}}
\def\vy{{\bm{y}}}
\def\vz{{\bm{z}}}

\def\mH{{\bm{H}}}
\def\mI{{\bm{I}}}
\def\mJ{{\bm{J}}}

\DeclareMathAlphabet{\mathsfit}{\encodingdefault}{\sfdefault}{m}{sl}
\SetMathAlphabet{\mathsfit}{bold}{\encodingdefault}{\sfdefault}{bx}{n}

\def\sR{{\mathbb{R}}}

\newcommand{\E}{\mathbb{E}}

\newcommand{\vepsilon}{\boldsymbol{\varepsilon}}
\usepackage{hyperref}
\usepackage{url}
\usepackage{physics}

\newtheorem{proposition}{Proposition}
\newtheorem{definition}{Definition}
\newtheorem{theorem}{Theorem}
\newtheorem{remark}{Remark}
\newtheorem{assumption}{Assumption}

\usepackage{algorithm}
\usepackage{algorithmic}
\usepackage{booktabs}   
\usepackage{longtable}
\usepackage{makecell}
\usepackage{multirow} 
\usepackage{tcolorbox}
\usepackage{textcomp}
\def\paperID{8198} 
\def\confName{ICCV}
\def\confYear{2025}

\title{Unleashing High-Quality Image Generation in Diffusion Sampling Using Second-Order Levenberg-Marquardt-Langevin}

\author{
    Fangyikang Wang$^{1*\ddagger}$\quad
    Hubery Yin$^{2*}$\quad
    Lei Qian$^{1}$\quad
    Yinan Li$^{1}$\quad
    Shaobin Zhuang$^{3\ddagger}$\quad
    Huminhao Zhu$^{1}$\quad\\
    Yilin Zhang$^{1}$\quad
    Yanlong Tang$^{4}$\quad
    Chao Zhang$^{1\dagger}$\quad
    Hanbin Zhao$^{1}$\quad
    Hui Qian$^{1}$\quad
    Chen Li$^{2}$\quad
    \\
    \footnotesize{$^1$Zhejiang University
    \quad $^2$WeChat Vision, Tencent Inc \quad} 
    \footnotesize{$^3$Shanghai Jiao Tong University \quad} 
    \footnotesize{$^4$Tencent Lightspeed Studio  \quad} \\
}

\begin{document}
\maketitle
\renewcommand{\thefootnote}{*}\footnotetext{Equal contribution \textsuperscript{$\dagger$} Corresponding author}
\renewcommand{\thefootnote}{$\ddagger$}\footnotetext{Work done as interns at WeChat Vision}
\begin{abstract}
The emerging diffusion models (DMs) have demonstrated the remarkable capability of generating images via learning the noised score function of data distribution.
Current DM sampling techniques typically rely on first-order Langevin dynamics at each noise level, with efforts concentrated on refining inter-level denoising strategies.
While leveraging additional second-order Hessian geometry to enhance the sampling quality of Langevin is a common practice in Markov chain Monte Carlo (MCMC), the naive attempts to utilize Hessian geometry in high-dimensional DMs lead to quadratic-complexity computational costs, rendering them non-scalable.
In this work, we introduce a novel Levenberg-Marquardt-Langevin (LML) method that approximates the diffusion Hessian geometry in a training-free manner, drawing inspiration from the celebrated Levenberg-Marquardt optimization algorithm.
Our approach introduces two key innovations: (1) A low-rank approximation of the diffusion Hessian, leveraging the DMs' inherent structure and circumventing explicit quadratic-complexity computations; (2) A damping mechanism to stabilize the approximated Hessian.
This LML approximated Hessian geometry enables the diffusion sampling to execute more accurate steps and improve the image generation quality.
We further conduct a theoretical analysis to substantiate the approximation error bound of low-rank approximation and the convergence property of the damping mechanism. 
Extensive experiments across multiple pretrained DMs validate that the LML method significantly improves image generation quality, with negligible computational overhead.
\end{abstract}
\begin{figure*}[!h]
\centering
\includegraphics[width=0.99\textwidth]{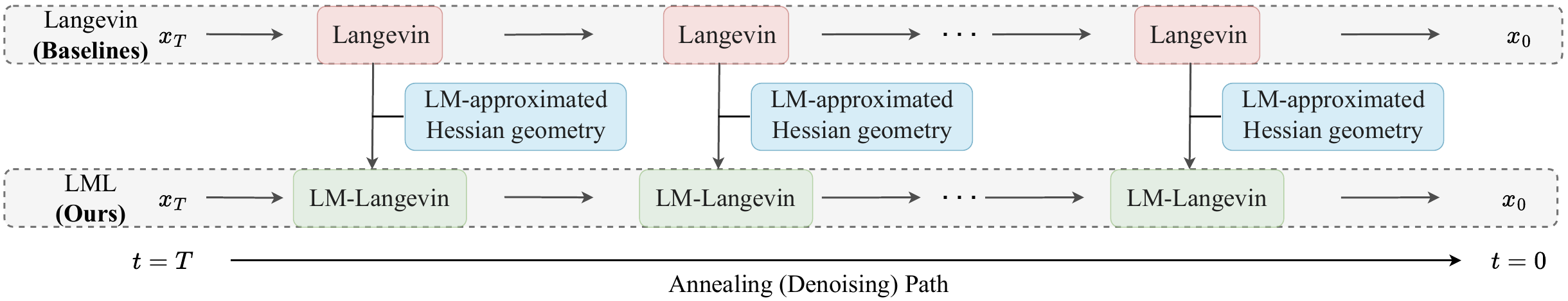} 
\vspace{-3mm}
\caption{Schematic comparison between our LML method with baselines. While previous works mainly focus on intriguing designs on the annealing path to improve diffusion sampling, they leave operations at specific noise levels to be first-order Langevin. Our approach proposes to leverage the Levenberg-Marquardt approximated Hessian geometry to guide the Langevin update to be more accurate.
}
\label{fig:anneal}
\vspace{-3mm}
\end{figure*}

\section{Introduction}
\label{sec:intro}

The emerging diffusion models (DMs) \cite{sohl2015deep, ho2020denoising,song2019generative, song2020score}, generating samples of data distribution from initial noise, have been proven to be an effective technique for modeling complex distribution, especially in generating high-quality images \cite{pmlr-v162-nichol22a, NEURIPS2021_49ad23d1, NEURIPS2022_ec795aea, ramesh2022hierarchical, rombach2022high, JMLR:v23:21-0635}. 
Training DMs can be viewed as using a neural network to match the (Stein) score function of the target distribution corrupted by different levels of noise. The sampling of DMs can be seen as running Langevin dynamics \cite{robert1999monte} using the learned diffused score and simultaneously slowly decreasing the noise level, which is called annealed Langevin dynamics \citep{song2019generative}.

Recently, many sampling methods have been proposed to enhance the quality of generated samples of pretrained DMs.
The majority of these efforts focus on refining the denoising scheme, as seen in works such as DDIM \cite{song2021denoising}, DPM-Solvers \cite{lu2022dpm,lu2022dpm++}, PNDM \citep{liu2022pseudo}, and others \citep{zhang2022fast,zheng2023dpm,zhao2024unipc,chen2024trajectory,zhou2024fast}.
Some also propose to select more critical denoising time-steps to enhance sampling quality \cite{jolicoeur2021gotta,gao2023fast,xue2024accelerating,xia2024towards}.
These efforts are primarily based on the analysis along the denoising path, yet they still perform first-order Langevin using the learned score within a specific noise level.

In the Langevin sampling area, it is common to employ additional second-order Hessian geometry to enhance the sampling quality. The additional second-order information can guide Langevin dynamics to take more accurate steps \cite{martin2012stochastic,chewi2020exponential}, which is called Newton-Langevin \cite{simsekli2016stochastic}.
A natural idea is that we can also leverage the Hessian geometry of DMs within each noise level to enhance the quality of diffusion sampling results.
However, it is extremely challenging to calculate the Hessian geometry within the context of high-dimensional DMs \cite{pmlr-v162-bao22d}, cause it requires quadratic-complexity computations. 
Current methods on diffusion Hessian either require auxiliary networks \cite{dockhorn2022genie,wang2025efficientlyaccessdiffusionfisher} or maintain memory-intensive states \cite{rissanen2025free}, making them inefficient for enhancing DM sampling.

In this paper, we introduce a novel method for approximating the diffusion Hessian within a specific noise level. 
Notably, our approach is the first of its kind to approximate the diffusion Hessian and apply it to enhance the sampling quality of pretrained, commercial-level DMs.
Our method, inspired by the celebrated Levenberg-Marquardt method in optimization \cite{levenberg1944method,marquardt1963algorithm,roweis1996levenberg}, is referred to as the Levenberg-Marquardt-Langevin (LML) method.
Our approach introduces two key innovations:
Initially, we derive a computationally tractable low-rank approximation of the diffusion Hessian by emulating the Gauss-Newton transformation, avoiding explicit quadratic-complexity computations. 
This approach capitalizes on the inherent structure of DMs to construct a low-rank approximation that captures critical geometric information.
Following this, we stabilize the approximated Hessian using the damping mechanism, which addresses its ill-conditioning, and acquire its inverse for geometric guidance.
Our method uses this LML-approximated Hessian geometry to enable the Langevin dynamics to take more accurate steps and improve image generation quality.

We also conduct comprehensive theoretical analyses for our low-rank approximation and damping mechanism. 
First, we establish the error bound for the low-rank approximation of the diffusion Hessian. Subsequently, we prove that the damping mechanism preserves an unbiased stationary measure and exhibit exponentially fast convergence in terms of $\chi^2$-divergence.

Extensive experiments on multiple pretrained DMs, including CIFAR-10, CelebA-HQ, SD-15, SD2-base, SD-XL and PixArt-$\alpha$, validate that the LML method significantly improves image generation quality with negligible computational overhead.

\begin{figure*}[!h]
\centering
\includegraphics[width=0.99\textwidth]{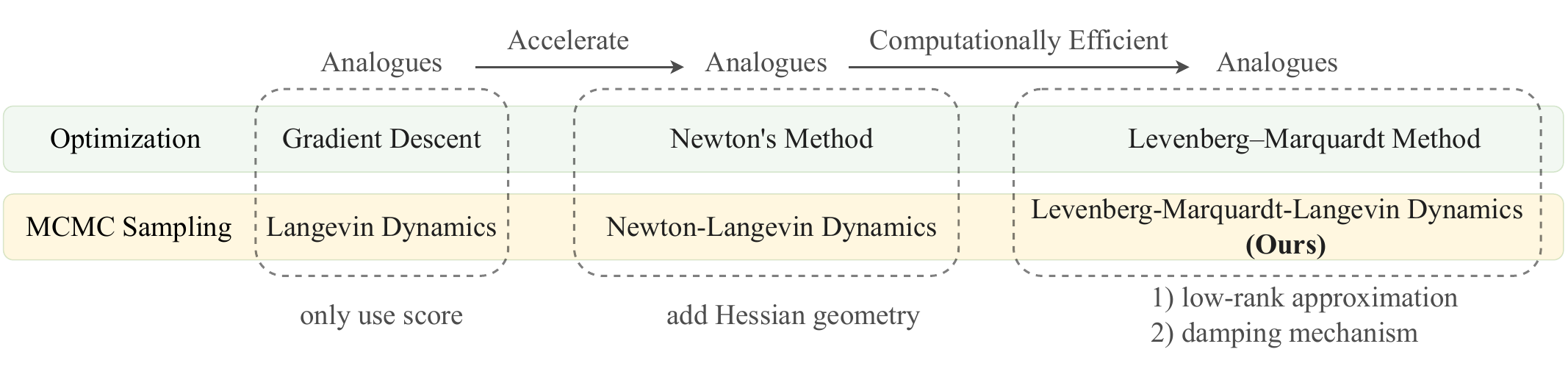} 
\vspace{-3mm}
\caption{The relation between optimization algorithms and MCMC sampling algorithms. We initially wanted to develop a diffusion sampler utilizing Hessian geometry, following the path of Newton-Langevin dynamics \cite{simsekli2016stochastic}.
However, this approach proved to be highly computationally expensive within the DM context.
Drawing inspiration from the Levenberg-Marquardt method used in optimization, our method incorporates low-rank approximation and damping techniques. This enables us to obtain the Hessian geometry in a computationally affordable manner. Subsequently, we use this approximated Hessian geometry to guide the Langevin updates.
}
\label{fig:}
\vspace{-3mm}
\end{figure*}
\section{Preliminaries}
\textbf{Notation}. The Euclidean norm over $\sR^{d}$ is denoted by $\norm{\cdot}$. Throughout, we simply write $\int g$ to denote the integral with respect to the Lebesgue measure: $\int g(x)\rd x$. 
When the integral is with respect to a different measure $\mu$, we explicitly write $\int g \rd \mu$. 
The expectation and variance of $g(X)$when $X \sim p$ are respectively denoted $\E_\mu g=\int g\rd\mu$ and $\text{var}_\mu g:=\int(g-\E_\mu g)^2\rd \mu$. 
When clear from context, we sometimes abuse notation by identifying a measure $\mu$ with its Lebesgue density.
We use $\mI_d$ to denote the $d$-dimensional identity matrix; when clear from context, we sometimes simply write $\mI$. $\mH_f$ and $\mJ_f$ denote the Hessian and Jacobian of $f$ respectively.

\subsection{Langevin Dynamics and Gradient Descent}
In statistics, the \textit{(Stein) score} of a distribution $p(\vx)$ is defined to be $\nabla_\vx \log p(\vx)$.
Given  an initial value $\vx_0 \sim \pi(\vx)$ with $\pi$ being a prior distribution,
Langevin dynamics (LD) can produce samples from  $p(\vx)$ using only the score function $\nabla_{\vx} \log p(\vx)$ following the SDE, 
\begin{equation} \label{ld_continuous}
\rd \vx_t= \nabla_{\vx} \log p\left(\vx_t\right) \rd t+\sqrt{2}\rd B_t,
\end{equation} 
where $B_t$ is the standard d-dimensional Brownian Motion (BM).
The distribution of $\vx_t$ equals $p(\vx)$ when 
$T \rightarrow \infty$, in which case $\vx_t$ becomes an exact sample from $p(\vx)$ under some regularity conditions \citep{welling2011bayesian}. 
Strictly speaking, a Metropolis-Hastings adjustment is needed, but it can often be ignored in practice  \citep{chen2014stochastic}. 
There is a deep connection involving the distribution of $\vx_t$ in LD to the renowned Gradient Descent method (GD) \citep{nesterov2018lectures} in optimization. 
The marginal distribution of a Langevin process $\left(\vx_t\right)_{t \geq 0}$ evolves according to a GD, over the Wasserstein probability space, that minimizes the Kullback-Leibler (KL) divergence $D_{\mathrm{KL}}(\cdot \|$ $\pi)$\citep{jordan1998variational,ambrosio2008gradient,villani2009optimal}.

\subsection{Diffusion Models and Annealed Langevin}

Suppose that we have a d-dimensional random variable $\vx(0) \in \sR^d$ following an unknown target distribution $p_0(\vx_0)$. Diffusion Models (DMs) define a forward process $\{\vx(t)\}_{t\in[0,T]}$ with $T>0$ starting with $\vx(0)$, such that the distribution of $\vx(t)$ conditioned on $\vx(0)$ satisfies
\begin{equation} \label{eq:diffusion_process}
  p_{t\vert 0}(\vx(t) \vert \vx(0)) = \mathcal{N}(\vx(t); \alpha(t) \vx(0), \sigma^2(t)\mathbf{I}), 
\end{equation}
where $\alpha(\cdot),\sigma(\cdot)\in \mathcal{C}(\left[ 0,T\right],\mathbb{R}^{+})$ have bounded derivatives, and we denote them as $\alpha_t$ and $\sigma_t$ for simplicity. The choice for $\alpha_t$ and $\sigma_t$ is referred to as the noise schedule of a DM.

DMs, both SMLD \citep{song2019generative} and DDPM \citep{ho2020denoising}, can be seen as learning a network to match the score of the diffused distribution $\log p_t(\vx_t)$ at different noise levels.
In practice, DMs usually use $\vepsilon_\theta(\vx(t),t)$ to estimate $-\sigma(t)\nabla_{\vx(t)} \log p_t(\vx(t), t)$ via optimizing the following denoising score matching objective:
\begin{eqnarray}\label{dpm_objective}
  \mathbb{E}_t \left\{ \lambda_t \mathbb{E}_{x_0, x_t} \left[ \lVert s_\theta(x_t, t) - \nabla_{x_t} \log p_(x_t, t | x_0, 0) \rVert^2 \right]\right\}.
\end{eqnarray}
The sampling process of diffusion models can be seen as annealed Langevin dynamics \cite{song2019generative}, which executes Langevin updates at each noise level while progressively reducing the noise scale.

\subsection{Langevin Guided by Hessian Geometry}
Newton's method (also called Newton–Raphson) \cite{polyak2007newton} is a famous optimization method that utilizes the second-order Hessian geometry to improve the GD. 
Developed as an analogy to Newton's method, the Newton-Langevin dynamics \citep{martin2012stochastic} utilize Hessian geometry to generate samples that adhere to the SDE
\begin{equation} \label{nld_continuous}
\rd \vx_t= \left[\nabla^2_{\vx}\log p\left(\vx_t\right) \right]^{-1}\nabla_{\vx} \log p\left(\vx_t\right) \rd t+\sqrt{2}\rd B'_t,
\end{equation} 
where the $B'_t$ is the BM scaled by the square-rooted Hessian geometry.
Calculating the Hessian geometry of high-dimensional distributions poses a significant challenge. Some studies have attempted to alleviate this issue through the Quasi-Newton technique \citep{simsekli2016stochastic,fu2016quasi}. However, these approaches fail to address the problem in the context of DM-scale dimensional data. For instance, the Stable Diffusion-v1.5 model \citep{rombach2022high} features a latent dimension of $16384$, resulting in a Hessian matrix of $16384 \times 16384$. 
Current methods on diffusion Hessian either require auxiliary networks training \cite{dockhorn2022genie,wang2025efficientlyaccessdiffusionfisher} or maintain memory-intensive states \cite{rissanen2025free}.
To the best of our knowledge, there is currently no effective method available to access the diffusion Hessian of advanced commercial-level DMs like SD-XL.

\subsection{Levenberg-Marquardt Method}
In the field of optimization, the Levenberg-Marquardt (LM) method \citep{roweis1996levenberg} is proposed as a computationally friendly and stabilized analogue to Newton's method.
Specifically, it modifies the computation of Hessian geometry in Newton's method in two key aspects: 
\begin{itemize}
\item \textbf{Low-rank approximated Hessian}
In the context of least squares problems, the Levenberg-Marquardt method often constructs a low-rank estimation of the Hessian geometry. Specifically, when $f(\vx) = \sum_{i=1}^m r_i(\vx)^2$, the Levenberg-Marquardt method approximates the Hessian into the following form, which is constructed from the Jacobians $\mJ_f(\vx)$.
\begin{equation} 
\label{low_rank_least_square} 
\mH_f(\vx) \approx 2\mJ_f(\vx)^\top\mJ_f(\vx).
\end{equation} 
This low-rank approximation technique is also referred to as the Gauss-Newton method in some literature \cite{eade2013gauss}. However, it is still unclear how to obtain a Levenberg-Marquardt-type low-rank estimation of the Hessian in the context of diffusion models.

\item \textbf{Damping mechanism}
The Levenberg-Marquardt Method introduces an additional damping identity matrix to the Hessian geometry. That is, it replaces the pure Hessian geometry $\left[\mH_f(\vx_k) \right]^{-1}$ in Newton's method with a combination expressed as $\left[\mH_f(\vx_k)+\lambda \mI \right]^{-1}$. Consequently, the Levenberg-Marquardt Method with damping is expressed as:
\begin{equation} 
  \vx_{k+1} = \vx_{k} - \eta\left[\mH_f(\vx_k)+\lambda \mI \right]^{-1}\nabla_\vx f (\vx_k).
\end{equation}
Empirical evidence suggests that the damping mechanism contributes to numerical stabilization \cite{kawamoto2009stabilization}, since $\lambda \mI$ can resolve the ill-conditioning of the Hessian. While the damping mechanism can be perceived as a trust region approach, it is more intuitive to view $\mH+\lambda \mI $ as a geometrical interpolation between $\mH$ and $\mI$. 
\end{itemize}

\begin{figure}[t]
\centering
\includegraphics[width=\columnwidth]{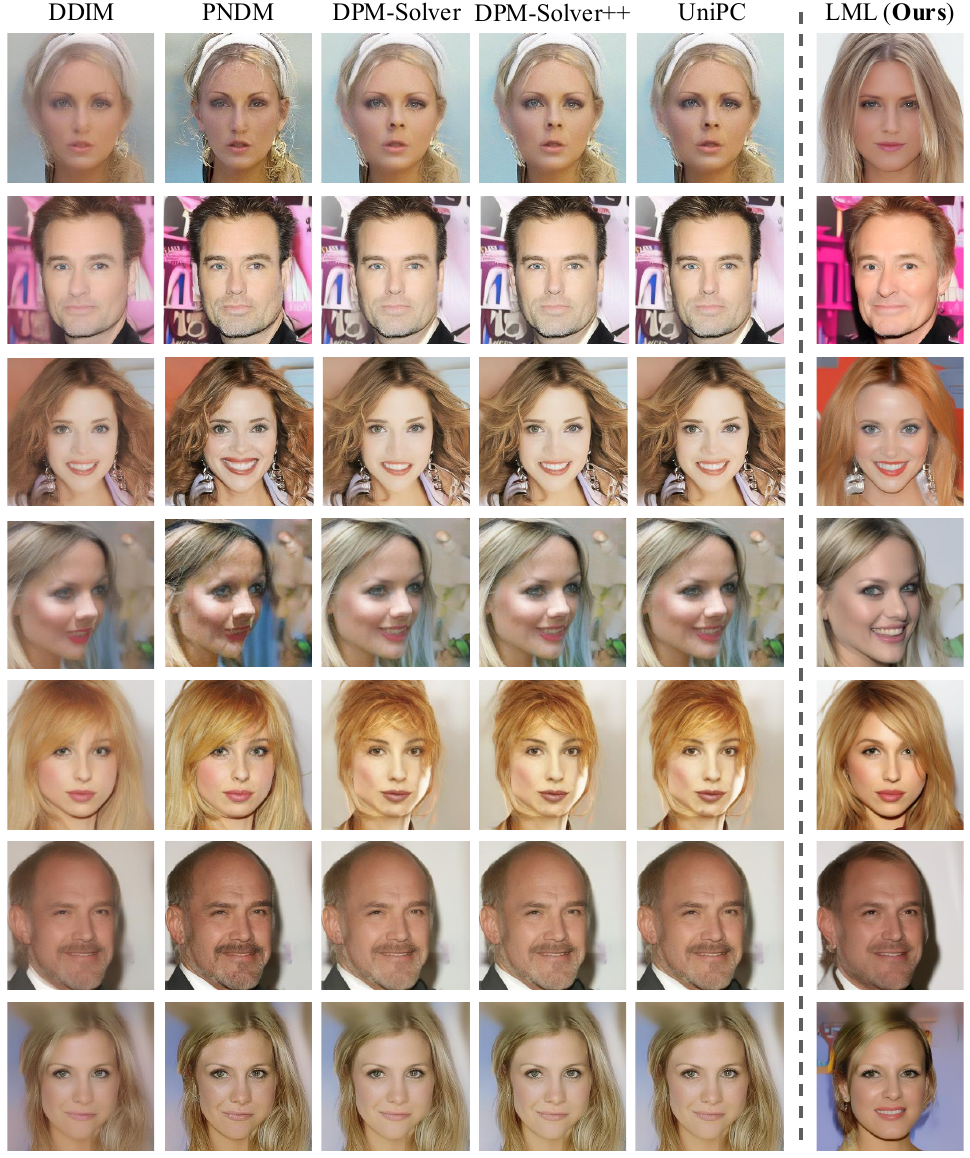} 
\caption{\textbf{Visual comparison} of images generated by our LML method and other methods, using the pre-trained LDM on CelebA-HQ 256$\cross$256 \cite{karras2017progressive} with the same seeds. It shows that our LML method contributes to more vivid and detailed generated images.
}
\label{fig:celeb-visual}
\vspace{-3mm}
\end{figure}

\section{Methodology}


\subsection{Low-rank Approximation of Diffusion Hessian}

Inspired by the low-rank estimation technique of the Hessian in the Levenberg-Marquardt method, we derive a similar low-rank estimation for the diffusion Hessian.
Specifically, we follow the intuition of the Levenberg-Marquardt method to approximate the diffusion Hessian by simplifying the second-order partial derivatives.
The low-rank approximate Hessian of diffusion models that we obtained is shown below. It is important to note that it includes noise schedule related coefficients, which distinguishes it from the least squares problem case in Eq. \ref{low_rank_least_square}.
\begin{proposition}\label{prop:gn_appro}
    Let $p_t$ be the diffused marginal distribution at time $t$ of the diffusion process, Eq. \ref{eq:diffusion_process} and $\vepsilon_\theta$ is learned via Eq. \ref{dpm_objective}, then the Hessian derivative of its log-density function $\nabla_{\vx_t}^2\log p_t(\vx_t)$ has a low-rank approximation form of $\frac{1}{\sigma(t)^2\norm{\vepsilon_\theta}^2}\vepsilon_\theta\vepsilon_\theta^{\top}$.
\end{proposition}
Notice that this approximation form is a scaled outer-product of the score.  The detailed derivation can be found in the Supplementary \ref{supp:gn_proof}.
For here and for the remainder of the paper, we will denote $\vepsilon_\theta(\vx(t),t)$ as $\vepsilon_\theta$, provided there is no risk of ambiguity.
In section \ref{sec:theo_error_bound}, we will show that this low-rank approximation possesses an error bound, ensuring that our approximation does not generate excessive errors.

\subsection{Damping Mechanism}
Similar to the scenario in the Levenberg-Marquardt method, we have obtained an accessible approximation of the diffusion Hessian as presented in Proposition \ref{prop:gn_appro}. 
However, this form is significantly ill-conditioned, making its inverse impossible to calculate. Specifically, the outer-product Hessian in Proposition \ref{prop:gn_appro} has an infinite condition number \cite{horn2012matrix}; hence, its inverse strictly does not exist in mathematical terms.
To address this issue, we propose adopting the damping mechanism used in the LML method. We introduce an additional damping identity matrix to the approximate Hessian as shown in Proposition \ref{prop:gn_appro}. 
This damped Hessian geometry is used in place of the full Hessian in the Newton Langevin in Eq. \ref{nld_continuous}, we consequently derive the following damping dynamics:
\begin{equation} \label{lml_continuous}
\rd \vx_t= \left[\nabla^2_{\vx}\log p\left(\vx_t\right) + \lambda \mI \right]^{-1}\nabla_{\vx} \log p\left(\vx_t\right) \rd t+\sqrt{2}\rd B'_t,
\end{equation} 
where the $B'_t$ is the BM scaled by the square-rooted LM approximated Hessian geometry.
Drawing inspiration from the Levenberg-Marquardt method, we refer to this damping Hessian Langevin in Eq. \ref{lml_continuous} as the Levenberg-Marquardt-Langevin dynamics (LML dynamics).
Similar to the Levenberg-Marquardt method, the damping coefficient $\lambda$ in Eq. \ref{lml_continuous} can be interpreted as an interpolation coefficient between the Hessian geometry and Identity geometry. Specifically, as $\lambda \to \infty$, Eq. \ref{lml_continuous} would degenerate to the standard Langevin with normalization on the geometry.
We observe that by adding the Levenberg-Marquardt damping identity matrix to the outer-product form of the Hessian in Equation Proposition \ref{prop:gn_appro}, its inverse can be conveniently computed using the Sherman-Morrison formula \citep{sherman1950adjustment}. 

\begin{equation} \label{sm-formula}
\begin{aligned}
\mH_{LM}^{-1}(\vx_t,\lambda)=&\left[\nabla^2_{\vx}\log p\left(\vx_t\right) + \lambda \mI \right]^{-1}\\
\approx&\left[\frac{1}{\sigma(t)^2\norm{\vepsilon_\theta}^2}\vepsilon_\theta\vepsilon_\theta^{\top} + \lambda \mI \right]^{-1}\\
=&\frac{1}{\sigma(t)^2\norm{\vepsilon_\theta}^2}\left[\vepsilon_\theta\vepsilon_\theta^{\top}+\lambda'\mI\right]^{-1}\\
=&\frac{1}{\lambda'\sigma(t)^2\norm{\vepsilon_\theta}^2}\left[\mI - \frac{\vepsilon_\theta\vepsilon_\theta^{\top}}{\lambda'+\norm{\vepsilon_\theta}^2}\right],
\end{aligned}
\end{equation} 
where $\lambda'=\sigma(t)^2\norm{\vepsilon_\theta}^2\lambda$. We also point out that the coefficients $\frac{1}{\lambda'\sigma(t)^2\norm{\vepsilon_\theta}^2}$ will be absorbed in the discretizing stepsize of the LML dynamics and only affect the norm.
The geometry direction essence is in the $\left[\mI - \frac{\vepsilon_\theta\vepsilon_\theta^{\top}}{\lambda'+\norm{\vepsilon_\theta}^2}\right]$ part. 
In the remainder of this paper, we will slightly abuse notation by not distinguishing between $\lambda$ and $\lambda'$. This simplification should not lead to any confusion.

\begin{algorithm}[t]
    \caption{Levenberg-Marquardt-Langevin (LML) diffusion sampler}
    \label{algo:lml-ds}
    \begin{algorithmic}[1] 
    \STATE \textbf{Input}: pretrained diffusion model noise predictor $\vepsilon_\theta$, number of timesteps $N$, noise schedule $\{\alpha_t\}$ and $\{\sigma_t\}$, Levenberg-Marquardt damping coefficient $\lambda>0$, EMA coefficient $\kappa$.
    \STATE Initiate $\vx_{\text{list}}=[]$, $\widetilde{\mH}_{N+1}^{-1} = \mI$
    \STATE Sample $\vx_N \sim \mathcal{N}(0,\sigma_{t_N} I)$.
    \FOR{$i = N, N-1,...,1$}
    \STATE $\vepsilon_i = \vepsilon_\theta(\vx_i,i)$
    \IF{$i\neq N$}
    \STATE $\widetilde{\vepsilon}_i = \kappa *\vepsilon_{i+1} +(1-\kappa)* \vepsilon_i$ \label{step:mix}
    \ELSE
    \STATE $\widetilde{\vepsilon}_i =  \vepsilon_i$
    \ENDIF
    \STATE $\widetilde{\mH}_i^{-1} = \mI - \frac{\widetilde{\vepsilon}_i\widetilde{\vepsilon}_i^{\top}}{\lambda+\norm{\widetilde{\vepsilon}_i}^2}$ \hfill\COMMENT{Levenberg-Marquardt approximate Hessian geometry}\label{step:lm-low-rank}
    \STATE $\vepsilon_i^{LM} = \widetilde{\mH}_i^{-1} \vepsilon_i$ 
    \hfill\COMMENT{Apply the approximate Hessian geometry}\label{step:apply-hessian}
    \STATE $\vepsilon_i^{LM}=\frac{\norm{\vepsilon_i}}{\norm{\vepsilon_i^{LM}}}\vepsilon_i^{LM}$ \hfill\COMMENT{Geometrical normalization}\label{step:normalization}
    \STATE $\vx_{i-1} = \text{DPM-Solver}(\vx_{\text{list}},\vepsilon_i^{LM},i)$ \label{step:next_state}
    \STATE $\vx_{\text{list}}=\vx_{\text{list}}\text{.append}(\vx_{i-1})$
    \ENDFOR
    \STATE \textbf{Output}: $\vx_0$
    \end{algorithmic}
\end{algorithm}

\begin{figure}[t]
\centering
\includegraphics[width=0.49\textwidth]{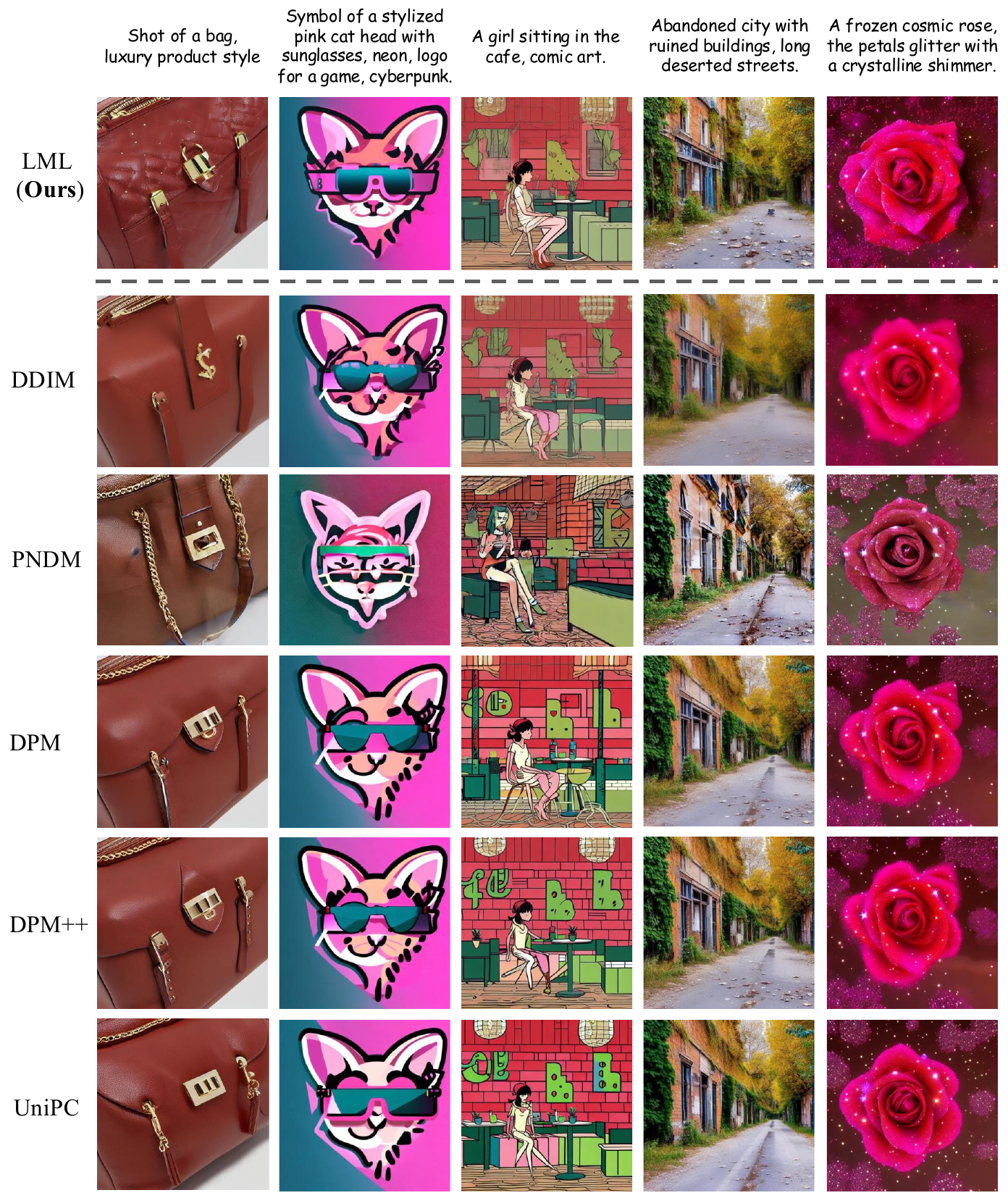} 
\caption{\textbf{Qualitative comparison} of our LML method and other methods in text-guided image generation.
The evaluation was performed on SD-15 \cite{rombach2022high}, using 10 NFEs and the same seeds.
}
\label{fig:sd_qualitative_sd15}
\vspace{-3mm}
\end{figure}

\subsection{Annealed Levenberg-Marquardt-Langevin}
To utilize the LML for diffusion sampling, we perform the discretization of Eq. \ref{lml_continuous} at each nose level, and gradually decrease the noise level.
We called this annealed LML, and point out that its continuous-time form would converge to the following SDE:
\begin{equation} \label{lm_reverse_sde}
  \rd x_t = \left[f_t \vx_t - g_t^2 \mH_{LM}^{-1} \nabla_x \log p_t(\vx_t)\right] \rd t + g_t \rd B'_t.
\end{equation}
Similar to the connection of the reverse SDE and diffusion ODE \cite{song2020score}, the LM reverse SDE in Eq. \ref{lm_reverse_sde} also has an associated ODE, which is a deterministic process that shares the same single-time marginal distribution:
\begin{equation} \label{lm_reverse_ode}
  \rd x_t = \left[f_t \vx_t - \frac{1}{2}g_t^2 \mH_{LM}^{-1} \nabla_\vx \log p_t(\vx_t)\right] \rd t.
\end{equation}
The following section will develop a practical LML sampler based on this deterministic LML ODE. 
\begin{remark}
    Our LM process can be viewed as the Legendre dual of the classical diffusion process with a transform map of $\nabla\log p_t(\cdot)+\lambda\norm{\cdot}$. See discussions in Supp. \ref{supp:mirror}.
\end{remark}

\subsection{Levenberg-Marquardt-Langevin Sampler}

We implement a diffusion sampler based on the LML diffusion ODE in Eq. \ref{lm_reverse_ode}, because it is suggested that deterministic diffusion samplers are far more efficient than the stochastic ones \cite{karras2022elucidating}. 
Our approach is generally consistent with the practices outlined in \cite{song2019generative}. 
The scheme for our LML diffusion sampler is illustrated in Algorithm \ref{algo:lml-ds}. 
Primarily, our sampler substitutes the first-order Langevin dynamics with LML at each noise level.
At each noise level, we initially compute the LM low-rank approximated and damping Hessian geometry, $\widetilde{\mH}_i^{-1}$, as detailed in step \ref{step:lm-low-rank} of Algorithm \ref{algo:lml-ds}. 
We incorporate the network output from previous steps, as shown in step \ref{step:mix}, following studies suggesting that this mixture is closer to the underlying true score due to the manifold's curvature \cite{liu2022pseudo}.
We then apply this approximate Hessian geometry to our initial gradient, $\vepsilon_i$, to obtain the Hessian-guided gradient, $\vepsilon_i^{LM}$, as detailed in step \ref{step:apply-hessian}. 
In terms of the stepsize for our LML sampler, in contrast to some sophisticated ways of choosing the stepsize of second-order methods \cite{hackl1980efficient,ku2012dynamical}, we follow the approach of \cite{fischer2021unit} to ensure that the Hessian geometry has a unit spectrum, which can be easily implemented through a normalization operation on the Hessian guided gradient $\vepsilon_i^{LM}$ as illustrated in step \ref{step:normalization}. 
The rescaling of BM in Eq. \ref{lml_continuous} can also be absorbed into this normalization operation.
Once the Hessian-guided gradient, $\vepsilon_i^{LM}$, is obtained, we adopt the DPM-Solver as our denoising scheme to calculate the next state, $\vx_{i-1}$, as shown in step \ref{step:next_state}.

It is worth noting that our LML does not require additional training, components, or network access. Instead, we only need several additional tensor arithmetic operations.


\begin{figure}[t]
\centering
\includegraphics[width=0.49\textwidth]{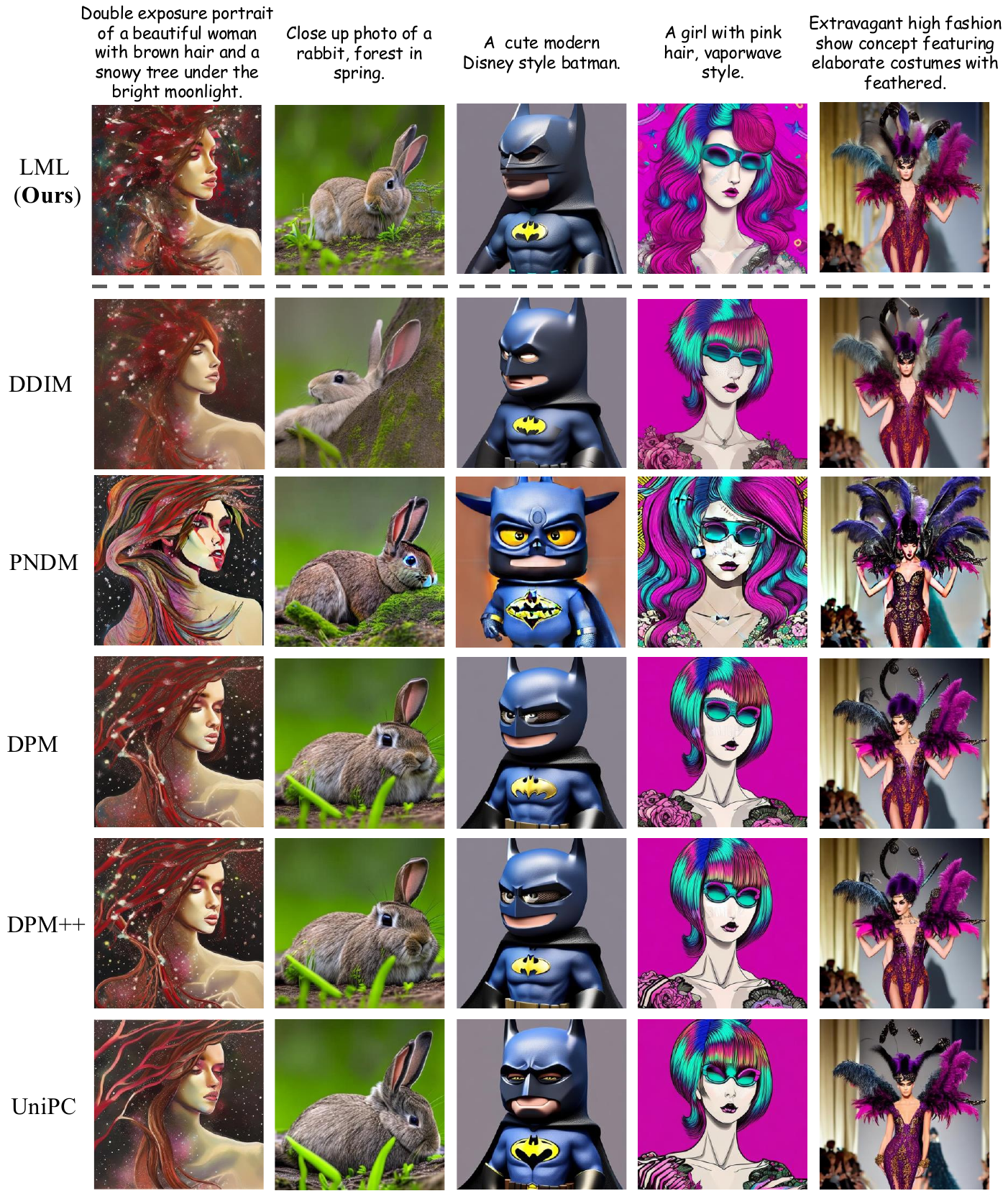} 
\caption{\textbf{Qualitative comparison} of our LML method and other methods in text-guided image generation.  The evaluation was performed on SD2-base \cite{rombach2022high}, using 10 NFEs and the same seeds.
}
\label{fig:sd_qualitative_sd2b}
\vspace{-3mm}
\end{figure}
\section{Theoretical Analysis}
In this section, we present rigorous theoretical analyses to substantiate the correctness and efficacy of our LML.

\subsection{Analysis on Low-rank Approximation}\label{sec:theo_error_bound}
Here, we establish the error bound of the low-rank approximation of the diffusion Hessian in Proposition \ref{prop:gn_appro}, thereby ensuring that our approximation does not introduce excessive errors. 
Given that we are evaluating the accuracy of a matrix-valued approximation, we choose to use the Hilbert-Schmidt norm \citep{gohberg1990hilbert} as a criterion.
\begin{proposition}\label{prop:error_bound_hessian}
    Assume that the norm of $\vx_t$ is bounded by $\delta_1$, the approximation error on $\vepsilon_\theta(\vx_t,t)$ is denoted as $\delta_2$, $\delta_3$ denote the bound on the second partial derivative of $\delta_1$ w.r.t. $\vx_t$, and $\mathcal{D}_y$ denote the dataset diameter.
    The approximation error of the LM low-rank Hessian, as referenced in Proposition \ref{prop:gn_appro}, is at most $\left(\delta_1+\alpha_t\delta_2+\alpha_t\mathcal{D}_y\right)\left(2+\delta_3+2\frac{\alpha_t^2}{\sigma_t^2}\mathcal{D}_y^2\right)$ when measured in terms of the Hilbert–Schmidt norm.
\end{proposition}
Our analysis relies on the analytical form diffusion Fisher \cite{wang2025efficientlyaccessdiffusionfisher}. The detailed proof can be found in Supp. \ref{supp:error_proof}.

\subsection{Analysis on Damping Mechanism}
Subsequently, we demonstrate the unbiased, exponentially fast convergence property of the damping mechanism.
\subsubsection{Stationary Measure}\label{sec:theo_stationary}
We confirm that the stationary measure of the damping dynamics, as defined in Eq. \ref{lml_continuous}, aligns with the target diffused distribution.
This ensures that the damping updates still guide samples towards the correct data distribution.

\begin{proposition}\label{prop:stationary}
Under mild regularity conditions, the stationary distribution of the damping dynamics in Eq. \ref{lml_continuous} exists and is unique, which also coincides with the marginal distribution $p_t\left(\vx_t\right)$ at every noise level.
\end{proposition}
We achieve this analysis by the Fokker-Planck equation \cite{risken1996fokker}. The detailed proof can be found in Supp. \ref{supp:stationary_proof}.

\begin{figure*}[t]
\centering
\begin{minipage}{0.33\textwidth}
\centering
\includegraphics[width=\linewidth]{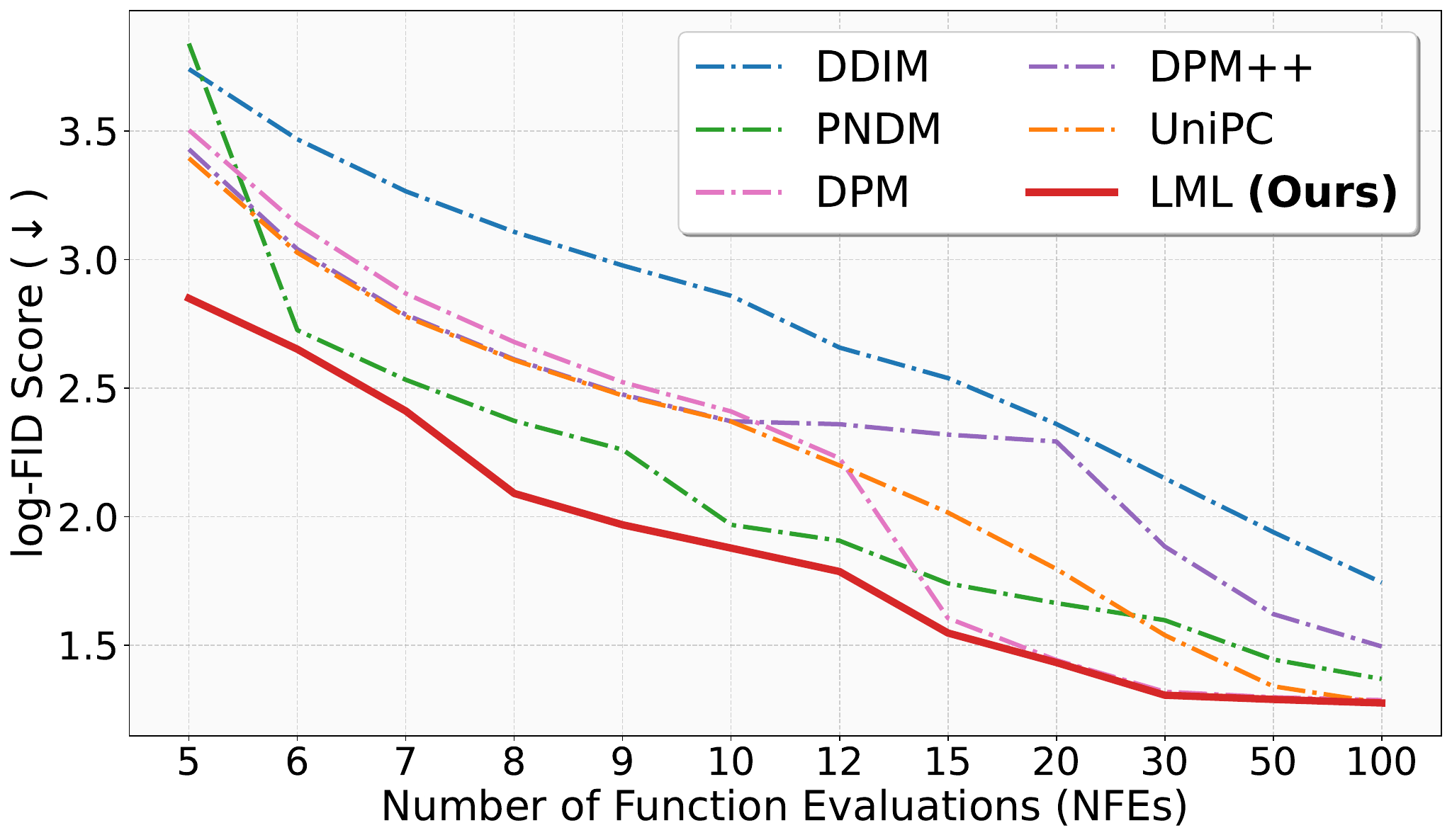}
\subcaption{CIFAR-10}\label{fig:cifar_line}
\end{minipage}
\hfill
\begin{minipage}{0.33\textwidth}
\centering
\includegraphics[width=\linewidth]{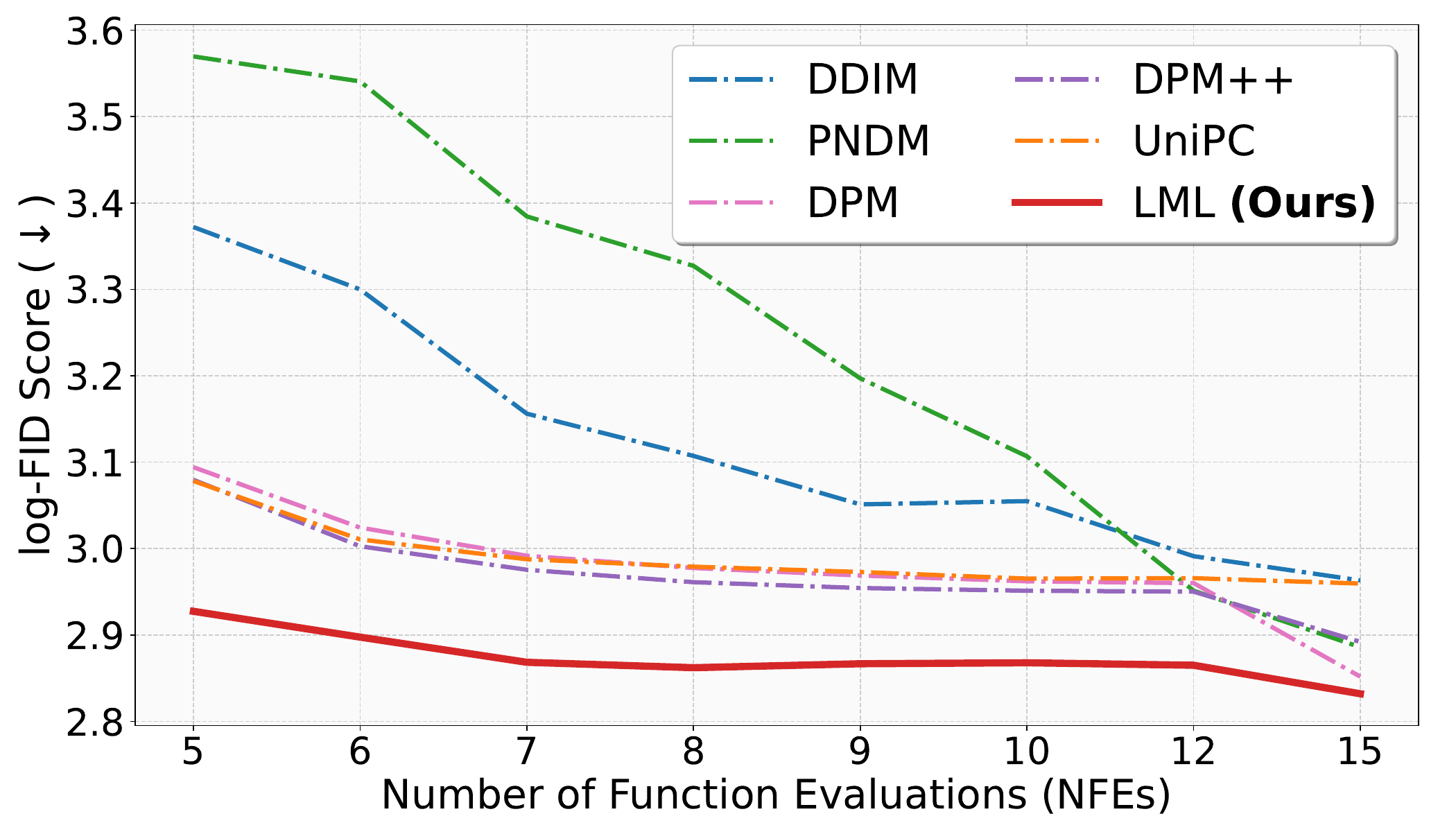}
\subcaption{SD-15 on MS-COCO}\label{fig:sd15_line}
\end{minipage}
\hfill
\begin{minipage}{0.33\textwidth}
\centering
\includegraphics[width=\linewidth]{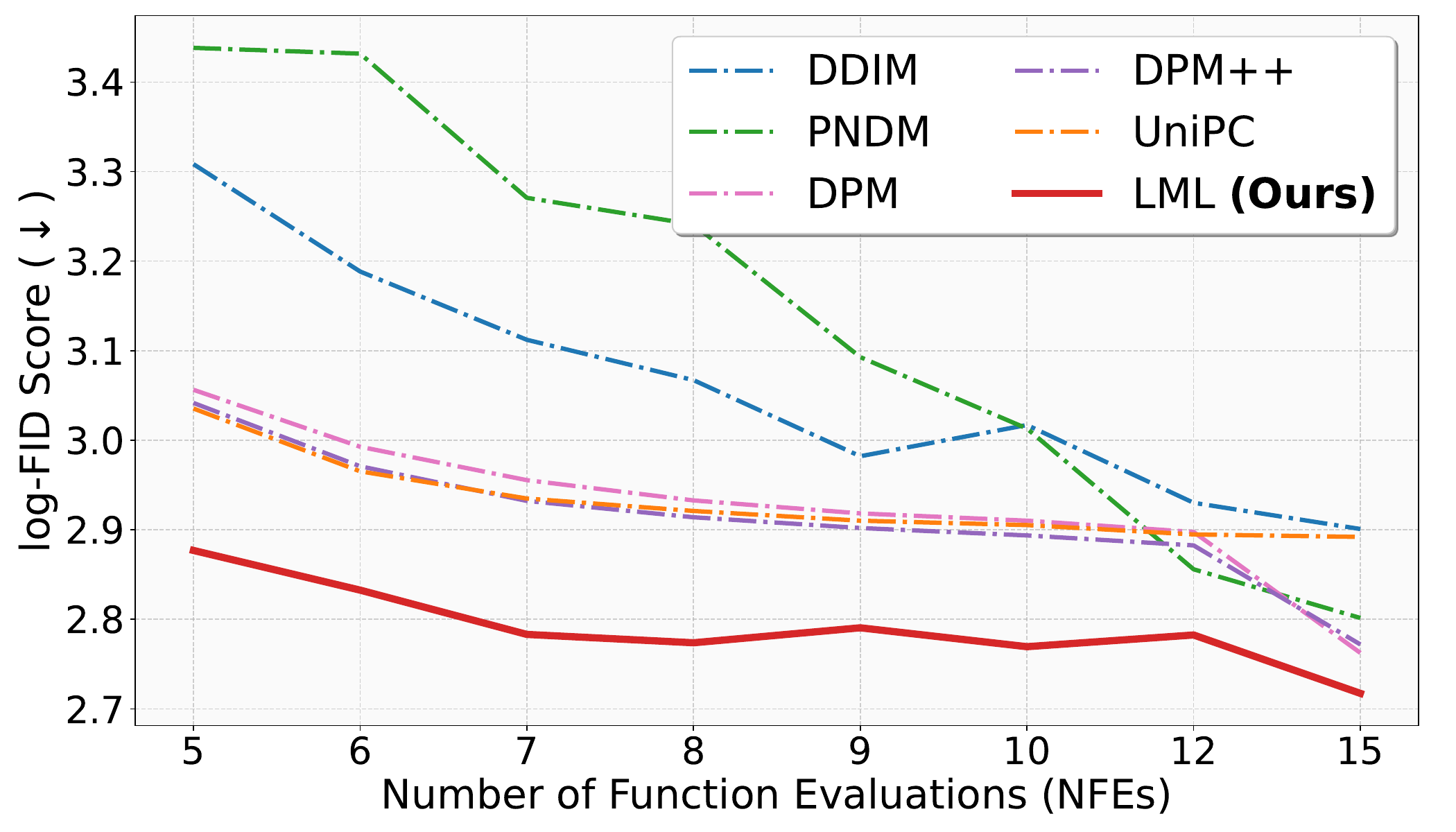}
\subcaption{SD2-base on MS-COCO}\label{fig:sd2b_line}
\end{minipage}
\caption{This line chart compares the log-FID scores ($\downarrow$) on (a) CIFAR-10, (b) MS-COCO with SD-15 and (c) MS-COCO with SD2-base.}
\end{figure*}

\subsubsection{Ergodic Convergence Rate}\label{sec:convergence}
We also want to determine the speed at which our damping dynamics converge. 
We demonstrate that our damping dynamics exhibit a satisfyingly fast, exponential convergence rate towards the target distribution.

\begin{proposition}\label{prop:lmd_convergence}
Let $\mu_t$ be the evolving distribution of the damping dynamics in Eq. \ref{lml_continuous}. We have that $\mu_t$ converges to the stationary distribution at an exponential ergodic convergence rate in terms of $\chi^2$-distance at every noise level, under certain regularity conditions.
\end{proposition}
We achieve this analysis by determining the relationship between Eq. \ref{lml_continuous} and the Mirror-Langevin \cite{chewi2020exponential}. The formal version and detailed proof can be found in Supp. \ref{supp:convergence_proof}.

\begin{remark}
    Once we establish the exponential ergodic convergence rate in terms of $\chi^2$-distance,
    The exponential convergence results for total variation distance \cite{verdu2014total}, Hellinger distance \cite{hellinger1909neue}, KL divergence \cite{kullback1951information} as illustrated in \citep[$\S 2.4$]{bickel2009springer}. And the exponential convergence results on Wasserstein distance \cite{villani2009optimal} can also be established with a log-Sobolev assumption, as illustrated in \citep{ding2015note}.
\end{remark}

\begin{table*}[t]
\centering
\renewcommand{\arraystretch}{1.1}
\caption{Comparison of different samplers on FID score( $\downarrow$) on CIFAR-10 unconditional generation. Best results are \textbf{bolded} and the second best results are \underline{underlined}. The FID scores were obtained by generating 50,000 samples, and all samplers were tested using the same seeds on the same checkpoint. It is shown that our LML always achieves the best or second best FID across all different NFEs.}
\resizebox{0.99\textwidth}{!}{
\begin{tabular}{c|cccccccccccc}
\Xhline{1pt}
{\multirow{2}{*}{Methods}} & \multicolumn{12}{c}{FID \textsubscript{($\downarrow$)} on CIFAR-10 generation} \\
\Xcline{2-13}{0.5pt}
& 5 NFEs & 6 NFEs & 7 NFEs & 8 NFEs & 9 NFEs & 10 NFEs & 12 NFEs & 15 NFEs & 20 NFEs & 30 NFEs & 50 NFEs & 100 NFEs \\
\Xhline{1pt}
DDIM \cite{song2021denoising} & 42.17& 32.09 & 26.21 & 22.38 & 19.64 & 17.45 & 14.27 & 12.67 & 10.60 & 8.58 & 6.96 & 5.72 \\
PNDM \cite{liu2022pseudo} &  46.56 & \underline{15.27} & \underline{12.59} & \underline{10.73} & \underline{9.59} & \underline{7.16} & \underline{6.73} & 5.70 & 5.28 &  4.94 & 4.24 & 3.93 \\
DPM-Solver \cite{lu2022dpm} &  33.26 & 23.06 & 17.61 & 14.58 & 12.47 & 11.13 & 9.28 & \underline{4.98} & \underline{4.23} & \underline{3.74} & \underline{3.66} & 3.62 \\
DPM-Solver++ \cite{lu2022dpm++} &  30.87 & 20.92 & 16.22 & 13.63 & 11.89 & 10.71 & 10.59 & 10.17 & 9.90 & 6.58 & 5.06 & 4.46 \\
UniPC \cite{zhao2024unipc} &  \underline{29.81} & 20.64 & 16.10 & 13.59 & 11.84 & 10.70 & 9.02 & 7.51 & 6.03 &  4.66 & 3.82 & \textbf{3.58} \\
\Xhline{0.2pt}
LML \textbf{(Ours)} &  \textbf{17.28} & \textbf{14.18} & \textbf{11.15} & \textbf{8.09} & \textbf{7.16} & \textbf{6.54} & \textbf{5.97} & \textbf{4.70} & \textbf{4.19} & \textbf{3.69} & \textbf{3.63} & \textbf{3.58} \\
\Xhline{1pt}
\end{tabular}}
\label{tab:unconditional_fid}
\end{table*}

\section{Experiments}
In this section, we validate the enhanced sampling quality of our LML method across various pretrained DMs. 
We evaluate in both pixel-space, latent-space, and text-guided conditional generation scenarios.
We select several most commonly-used and advanced sampling methods such as DDIM \cite{song2020score}, PNDM \cite{liu2022pseudo}, DPM-Solver \cite{lu2022dpm}, DPM-Solver++ \cite{lu2022dpm++} and UniPC \cite{zhao2024unipc} as baselines.
We will also demonstrate the computational efficiency of our LML method. 
All experiments are executed using open-source, pretrained DMs, with the datatype set to float32 and the timestep scheme as uniform.
More experimental details and results are deferred to Supp. \ref{supp:exp_detail} and \ref{supp:addition_results}.

\begin{table}[t]
    \centering
    \renewcommand{\arraystretch}{1.1}
    \caption{Comparison of different samplers on CelebA-HQ unconditional generation. Best results are \textbf{bolded} and the second best results are \underline{underlined}.}
    \resizebox{0.49\textwidth}{!}{%
        \begin{tabular}{c|c|cc|ccc}
            \Xhline{1pt}
            {\multirow{2}{*}{Methods}} & Colorful & \multicolumn{2}{c|}{Face Quality} & \multicolumn{3}{c}{Aesthetic }\\
            \Xcline{2-7}{0.5pt}
            & ColorS\textsubscript{($\uparrow$)} & {FS\textsubscript{($\uparrow$)}} & DFIQA\textsubscript{($\uparrow$)} & PicS\textsubscript{($\uparrow$)} & EAT\textsubscript{($\uparrow$)} & Laion\textsubscript{($\uparrow$)} \\
            \Xhline{1pt}
            DDIM \cite{song2021denoising} & 34.13 & 4.85 & 0.539 & 19.85 & \underline{4.61} & 5.24 \\
            PNDM \cite{liu2022pseudo}& \underline{38.29} & 4.89 & 0.553 & 19.70 & 4.28 & 5.11 \\
            DPM \cite{lu2022dpm} & 34.85 & 5.06 & 0.566 & \underline{20.00} & 4.51 & \underline{5.29} \\
            DPM++ \cite{lu2022dpm++} & 35.08 & 5.08 & 0.560 & 19.97 & 4.46 & 5.26 \\
            UniPC \cite{zhao2024unipc} & 35.95 & \underline{5.09} & \underline{0.568} & 19.97 & 4.47 & 5.26 \\
            \Xhline{0.2pt}
            LML\textbf{(Ours)} & \textbf{40.53} & \textbf{5.24} & \textbf{0.607}  &\textbf{20.98}  &  \textbf{4.75} & \textbf{5.37} \\
            \Xhline{1pt}
        \end{tabular}
    }
    \label{tab:celeba}
    \vspace{-3mm}
\end{table}

\subsection{Pixel-Space Image Generation}
We initially compare the unconditional sampling quality of our LML method with baselines on the CIFAR-10 dataset \cite{alex2009learning}.
For each sampler, we generate 50,000 samples for FID evaluation.
As illustrated in Table \ref{tab:unconditional_fid} and Figure \ref{fig:cifar_line}, our approach improves the sampling performance of the baseline Langevin methods in most NFE scenarios.

\subsection{Latent-Space Image Generation}
We evaluated our LML on the LDM \cite{rombach2022high} that was trained on CelebA-HQ \cite{karras2017progressive} at a resolution of 256×256. In Table \ref{tab:celeba}, we employed five metrics spanning three aspects to demonstrate the superiority of LML. These aspects include colorfulness \cite{hasler2003measuring}, face quality (FS \cite{liao2024facescore} and DFIQA \cite{chen2024dsl}), and human-preference-aesthetic scores (PicS \cite{kirstain2023pick}, EAT \cite{he2023eat} and Laion-Aes \cite{schuhmann2022laion}). Additionally, we present a visual comparison in Figure \ref{fig:celeb-visual}. All these experiments were carried out with a NFE of 10. The results were tested by averaging over 1000 samples and the same seeds.

\begin{table}[t]
    \centering
    \renewcommand{\arraystretch}{1.1}
        \caption{Comparison of FID score \textsubscript{($\downarrow$)} for the task of text-guided conditional generation of SD on randomly selected 30,000 MS-COCO prompts. Best results are \textbf{bolded} and the second best results are \underline{underlined}.}
    \resizebox{0.49\textwidth}{!}{
        \begin{tabular}{c|cccccccc}
            \Xhline{1pt}
            {\multirow{2}{*}{Methods \textbackslash NFEs}} & \multicolumn{8}{c}{FID \textsubscript{($\downarrow$)} on MS-COCO-14 prompts} \\
            \Xcline{2-9}{0.5pt}
            & 5 & 6 & 7 & 8 & 9 & 10 & 12 & 15 \\
            \Xhline{0.5pt}
            \Xcline{1-9}{0.5pt} & \multicolumn{8}{c}{SD-1.5} \\
            \Xhline{0.5pt}
            DDIM \cite{song2021denoising} & 29.14 & 27.11 & 23.48 & 22.36 & 21.14 & 21.22 & 19.91 & 19.36 \\
            PNDM \cite{liu2022pseudo} &  35.50	&34.49	&29.50&	27.86&	24.46&	22.35&	19.13	&17.92\\
            DPM \cite{lu2022dpm} & 22.07 & 20.58 & 19.92 & 19.64 & 19.47 & 19.34 & 19.30	&\underline{17.32}\\
            DPM++ \cite{lu2022dpm++} & 21.75&	\underline{20.14}	& \underline{19.60}	&\underline{19.32}	&\underline{19.19}	&\underline{19.13}	&\underline{19.11}	&18.03\\
            UniPC \cite{zhao2024unipc} & \underline{21.72}&	20.30	&19.84	&19.67	&19.55	&19.40	&19.41	&19.29\\
            \Xhline{0.2pt}
            LML \textbf{(Ours)} & \textbf{18.68} & \textbf{18.13} & \textbf{17.61} & \textbf{17.50} & \textbf{17.58} & \textbf{17.60} & \textbf{17.55} & \textbf{16.98} \\
            \Xhline{0.5pt}
            \Xcline{1-9}{0.5pt} & \multicolumn{8}{c}{SD2-base} \\
            \Xhline{0.5pt}
            DDIM \cite{song2021denoising} & 27.34 & 24.25 & 22.47 & 21.48 & 19.73 & 20.43 &  18.73& 18.19 \\
            PNDM \cite{liu2022pseudo} &  31.13	&30.93&	26.33	&25.56&	22.04	&20.35	&\underline{17.39}	&16.47 \\
            DPM \cite{lu2022dpm} & 21.25 & 19.94 & 19.21 & 18.78 & 18.51 & 18.36 & 18.13 & \underline{15.84} \\
            DPM++ \cite{lu2022dpm++} & 20.94&	19.51&	\underline{18.77}&	\underline{18.43}	&\underline{18.21}	&\underline{18.06}&	17.86&	15.99  \\
            UniPC \cite{zhao2024unipc} &  \underline{20.81}&	\underline{19.40}	&18.82&	18.56&	18.36&	18.27&	18.08	&18.03  \\
            \Xhline{0.2pt}
            LML \textbf{(Ours)} & \textbf{17.76} & \textbf{16.99} & \textbf{16.17} & \textbf{16.02} & \textbf{16.29} & \textbf{15.95} &  \textbf{16.16}& \textbf{15.14} \\
            \Xhline{1pt}
        \end{tabular}
    }
    \label{tab:fid_coco}
\end{table}

\begin{table*}[t]
    \centering
    \renewcommand{\arraystretch}{1.13}
    \caption{Comparison of different samplers in text-guided image generation task on the T2I-BC benchmark \cite{huang2023t2i}. The metrics are Color, Shape, and Texture. Best results are \textbf{bolded} and the second best results are \underline{underlined}.}\label{tab:t2i}
    \resizebox{0.99\textwidth}{!}{
        \begin{tabular}{c|ccc|ccc|ccc|ccc}
            \Xhline{1pt}
            \multirow{2}{*}{Metrics} 
            & \multicolumn{3}{c|}{T2I Benchmark on SD-1.5 \cite{rombach2022high}}
            & \multicolumn{3}{c|}{T2I Benchmark on SD2-base \cite{rombach2022high}}
            & \multicolumn{3}{c|}{T2I Benchmark on SD-XL \cite{podell2023sdxl}}
            & \multicolumn{3}{c}{T2I Benchmark on PixArt-$\alpha$ \cite{chen2023pixart} }\\
            \Xcline{2 - 13}{0.2pt} 
            & {Color}\textsubscript{($\uparrow$)} & Shape\textsubscript{($\uparrow$)} & Texture\textsubscript{($\uparrow$)}  
            & {Color}\textsubscript{($\uparrow$)} & Shape\textsubscript{($\uparrow$)} & Texture\textsubscript{($\uparrow$)} 
            & {Color}\textsubscript{($\uparrow$)} & Shape\textsubscript{($\uparrow$)} & Texture\textsubscript{($\uparrow$)}  
            & {Color}\textsubscript{($\uparrow$)} & Shape\textsubscript{($\uparrow$)} & Texture\textsubscript{($\uparrow$)} \\
            \Xhline{1pt}
            DDIM \cite{song2021denoising} & 0.3864 & 0.3791 & 0.4231 & 0.5111 & 0.4182 & 0.4822 & 0.5670 & 0.4712 & 0.5076 & 0.2933 & 0.3746 & 0.4045\\
            PNDM \cite{liu2022pseudo} & 0.3810 & 0.3761 & \underline{0.4331} & 0.5067 & 0.4190 & 0.4863 & 0.5682 & 0.4772 & 0.5085 & \textbf{0.4035} & \underline{0.3996} & 0.4225\\
            DPM \cite{lu2022dpm} & 0.3877 & 0.3945 & 0.4294 & 0.5162 & 0.4307 & 0.5002 & 0.5795 & 0.4841 & 0.5194 & 0.3263 & 0.3895 & 0.4349\\
            DPM++ \cite{lu2022dpm++} & 0.3881 & \underline{0.3958} & 0.4307 & \underline{0.5202} & 0.4326 & \underline{0.5020} & 0.5798 & 0.4858 & 0.5201 & 0.3136 & 0.3882 & \underline{0.4364}\\
            UniPC \cite{zhao2024unipc}  & \underline{0.3892} & 0.3889 & 0.4306 & 0.5146 & \underline{0.4337} & 0.4995 & \underline{0.5828} & \underline{0.4910} & \underline{0.5234} & 0.3187 & 0.3880 & 0.4266\\
            \Xhline{0.2pt}
            LML \textbf{(Ours)} & \textbf{0.4335} & \textbf{0.4252} & \textbf{0.4746} & \textbf{0.5640} & \textbf{0.4792} & \textbf{0.5330} & \textbf{0.5908} & \textbf{0.4938} & \textbf{0.5363} & \underline{0.3801} & \textbf{0.4265} & \textbf{0.4684}\\
            \Xhline{1pt}
        \end{tabular}
    }
\end{table*}

\subsection{Text-guided Image Generation}
\subsubsection{Qualitative Comparison}
Qualitative comparison on SD-15 and SD2-base generation is provided in Figure \ref{fig:sd_qualitative_sd15} and \ref{fig:sd_qualitative_sd2b}, clearly indicating that our LML method enhances the quality of image details under the same seed and prompt with a NFE of 10.

\subsubsection{FID on MS-COCO Benchmark}
Following the approach in \cite{saharia2022photorealistic,zhang2024tackling}, we randomly selected 30,000 prompts from the MS-COCO dataset \cite{lin2014microsoft} and generated images conditioned on these prompts.
We tested our sampler in terms of FID on the SD-1.5 and SD2-base models with resolutions of 512$\times$512 and a CFG scale 7.0.
Table \ref{tab:fid_coco} shows that our LML sampler surpasses the baseline methods on SD models across different NFEs.


\subsubsection{T2I-CB Benchmark}
To further validate the enhanced sampling quality of LML in text-to-image generation, we carried out tests on diverse commercial-level DMs (SD-1.5 \cite{rombach2022high}, SD2-base \cite{rombach2022high}, SD-XL \cite{podell2023sdxl}, and PixArt-$\alpha$ \cite{chen2023pixart}) using the T2I-CB \cite{huang2023t2i} benchmark.
This benchmark is designed for open-world text-to-image generation evaluation. For each model, we evaluated three metrics (color, shape, and texture) to assess the visual quality of the samplers. Each assessment was based on 30,000 images with a NFE of 10.
As shown in Table \ref{tab:t2i}, our LML always achieves the best or second best scores on T2I-BC across these models.

\subsubsection{Application on ControlNet}
\begin{figure}[t]
\centering
\includegraphics[width=\columnwidth]{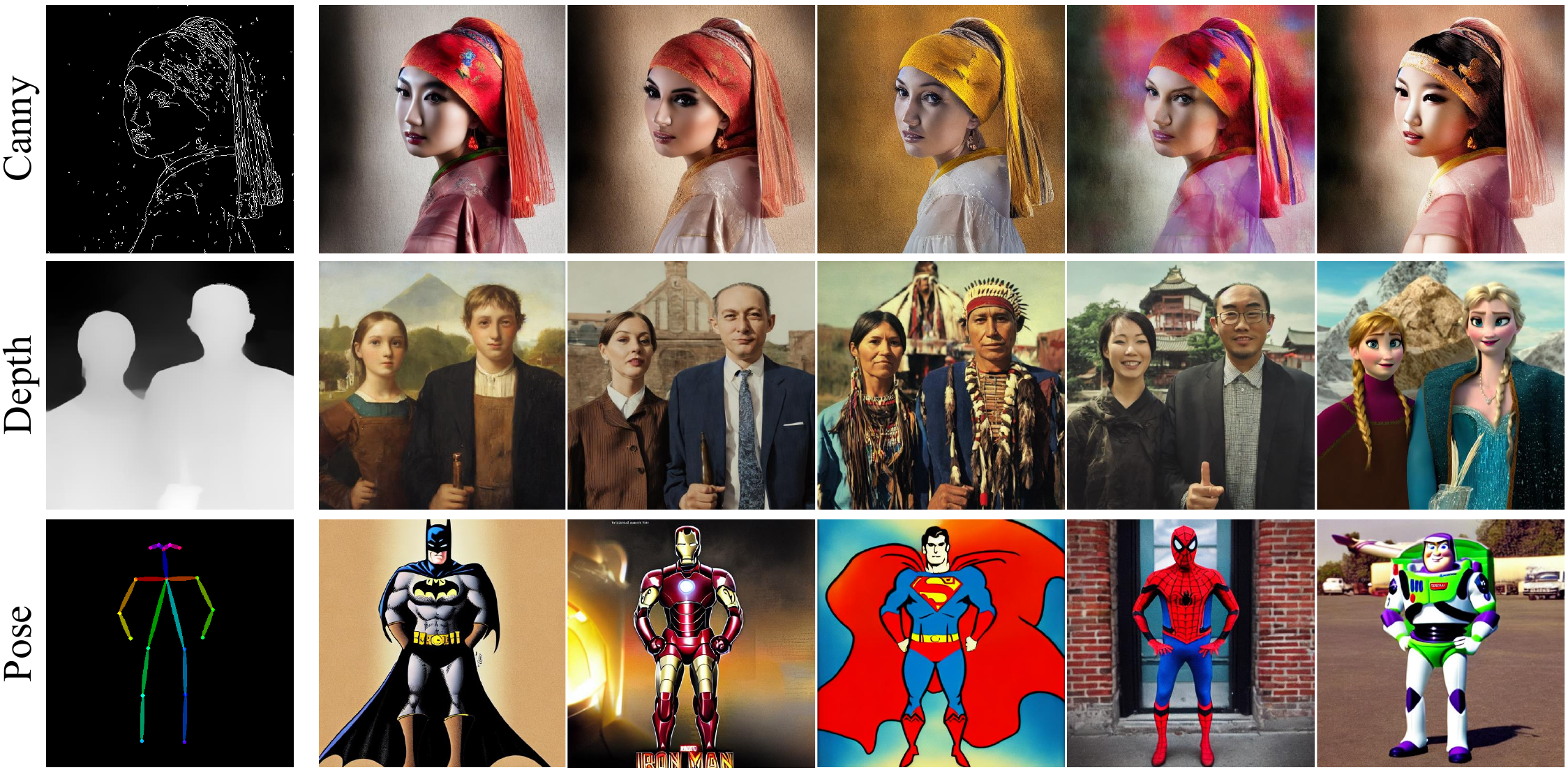} 
\caption{LML integrates seamlessly with ControlNet.
}
\label{fig:control-net}
\end{figure}

Our model can seamlessly integrate with existing diffusion model plugins, such as ControlNet \cite{zhang2023adding}.
The results in Fig \ref{fig:control-net} demonstrate that our method is fully compatible with ControlNet and generates high-quality samples.

\subsection{Time-costs Comparison}
Table \ref{tab:time_cost} presents the average wall-clock time costs for LML across different image generation tasks. 
These time-cost experiments were conducted on a single NVIDIA RTX 4090 chip and averaged over 200 tests with a batchsize of 1. 
The results suggest that the additional computational cost incurred by our LML technique is virtually negligible, making LML as time-efficient as DDIM \cite{song2021denoising}.

\begin{table}[t] 
\centering 
\renewcommand{\arraystretch}{1.1} 
\caption{Average wall-clock time costs for LML and DDIM.
} 
\resizebox{0.47\textwidth}{!}{
\begin{tabular}{c|c|ccccccc} 
\Xhline{1pt} 
\multirow{2}{*}{Models}& \multirow{2}{*}{Methods \textbackslash NFEs}& \multicolumn{7}{c}{Time-costs (s)}\\
\Xcline{3-9}{0.2pt} 
&  & 5 & 10 & 15 & 20 & 30 & 50 & 100 \\ 
\Xhline{1pt} 
\multirow{2}{*}{CIFAR-10}&DDIM & 0.13 & 0.20 & 0.27 & 0.34 & 0.52 & 0.81 & 1.49 \\ 
&LML \textbf{(Ours)} & 0.13 & 0.21 & 0.27 & 0.37 & 0.48 & 0.82 & 1.51 \\ 
\Xhline{0.5pt} 
\multirow{2}{*}{CelebA}&DDIM & 0.20 & 0.28 & 0.39 & 0.47 & 0.66 & 1.03 & 1.95 \\ 
&LML \textbf{(Ours)} & 0.20 & 0.31 & 0.38 & 0.48 & 0.67 & 1.10 & 2.08 \\ 
\Xhline{0.5pt} 
\multirow{2}{*}{SD-1.5}&DDIM & 0.45 & 0.73 & 1.03 & 1.31 & 1.87 & 3.01 & 5.85 \\ 
&LML \textbf{(Ours)} & 0.45 & 0.74 & 1.04 & 1.32 & 1.89 & 3.03 & 5.89 \\ 
\Xhline{0.5pt} 
\multirow{2}{*}{SD-2base}&DDIM & 0.45 & 0.71 & 1.00 & 1.25 & 1.80 & 2.87 & 5.53 \\ 
&LML \textbf{(Ours)} & 0.46 & 0.72 & 0.98 & 1.27 & 1.81 & 2.88 & 5.60 \\ 
\Xhline{0.5pt} 
\multirow{2}{*}{SD-XL}&DDIM & 2.32 & 4.25 & 6.23 & 8.12 & 12.06 & 19.82 & 39.48 \\ 
&LML \textbf{(Ours)} & 2.32 & 4.30 & 6.23 & 8.21 & 12.11 & 19.99 & 39.55 \\ 
\Xhline{0.5pt} 
\multirow{2}{*}{PixArt}&DDIM & 0.66 & 1.08 & 1.49 & 1.92 & 2.74 & 4.31 & 8.45 \\ 
&LML \textbf{(Ours)} & 0.67 & 1.08 & 1.51 & 1.92 & 2.75 & 4.28 & 8.49 \\ 
\Xhline{1pt} 
\end{tabular} 
}
\label{tab:time_cost} 
\end{table}

\subsection{Hyperparameters}
Our LML introduces two robust hyperparameters with clear high-level interpretations: the damping coefficient $\lambda$ and the mixture coefficient $\kappa$. 
The damping coefficient $\lambda$ determines the degree to which our approximated Hessian is interpolated with the identity matrix. 
A larger $\lambda$ value results in the dominance of the identity matrix, leading the LML to degenerate into Langevin. Conversely, a smaller $\lambda$ value allows more guidance from the Hessian geometry, but a minimal value could lead to ill-conditioning issues.
The $\kappa$ controls the EMA rate of previous information. A small $\kappa$ would make the LML close to the identity map, but LML contributes to obvious image quality enhancement in practice.
More discussions on hyperparameter tuning scheme, setup, and ablation experiments are provided in Supp. \ref{supp:hyperparameter}.

\section{Conclusions}
In this paper, we propose a training-free method termed Levenberg-Marquardt-Langevin (LML) for improving sampling quality, which uses a low-rank approximated damping Hessian geometry to guide the Langevin update.
We provide a theoretical analysis of the approximation error for low-rank approximation, the stationary measure, and the convergence rate for damping dynamics.
We conduct extensive experiments to demonstrate that our LML method can contribute to significant improvement in sampling quality, with negligible computational overhead. 
The code is available at \url{https://github.com/zituitui/LML-diffusion-sampler}.
\section*{Acknowledgments} 
This work was supported in part by the National Natural Science Foundation of China under Grants 62206248 and 62402430, and the Zhejiang Provincial Natural Science Foundation of China under Grant LQN25F020008.
Fangyikang Wang would like to extend his gratitude to Pengze Zhang from ByteDance, Binxin Yang, and Xinhang Leng from WeChat Vision for their discussions regarding the experiments. Additionally, he is thankful to Zebang Shen from ETH Zürich and Zhichao Chen from Peking University for their insights on Langevin dynamics.

{
    \small
    \bibliographystyle{ieeenat_fullname}
    \bibliography{main}

\begin{thebibliography}{95}
\providecommand{\natexlab}[1]{#1}
\providecommand{\url}[1]{\texttt{#1}}
\expandafter\ifx\csname urlstyle\endcsname\relax
  \providecommand{\doi}[1]{doi: #1}\else
  \providecommand{\doi}{doi: \begingroup \urlstyle{rm}\Url}\fi

\bibitem[Alex(2009)]{alex2009learning}
Krizhevsky Alex.
\newblock Learning multiple layers of features from tiny images.
\newblock \emph{https://www. cs. toronto. edu/kriz/learning-features-2009-TR. pdf}, 2009.

\bibitem[Ambrosio et~al.(2008)Ambrosio, Gigli, and Savar{\'e}]{ambrosio2008gradient}
Luigi Ambrosio, Nicola Gigli, and Giuseppe Savar{\'e}.
\newblock \emph{Gradient flows: in metric spaces and in the space of probability measures}.
\newblock Springer Science \& Business Media, 2008.

\bibitem[Bao et~al.(2022)Bao, Li, Sun, Zhu, and Zhang]{pmlr-v162-bao22d}
Fan Bao, Chongxuan Li, Jiacheng Sun, Jun Zhu, and Bo Zhang.
\newblock Estimating the optimal covariance with imperfect mean in diffusion probabilistic models.
\newblock In \emph{Proceedings of the 39th International Conference on Machine Learning}, pages 1555--1584. PMLR, 2022.

\bibitem[Bickel et~al.(2009)Bickel, Diggle, Fienberg, Gather, Olkin, and Zeger]{bickel2009springer}
P Bickel, P Diggle, S Fienberg, U Gather, I Olkin, and S Zeger.
\newblock Springer series in statistics.
\newblock \emph{Principles and Theory for Data Mining and Machine Learning. Cham, Switzerland: Springer}, 2009.

\bibitem[Bris and Lions(2008)]{bris2008existence}
C~Le Bris and P-L Lions.
\newblock Existence and uniqueness of solutions to fokker--planck type equations with irregular coefficients.
\newblock \emph{Communications in Partial Differential Equations}, 33\penalty0 (7):\penalty0 1272--1317, 2008.

\bibitem[Byrd et~al.(1995)Byrd, Lu, Nocedal, and Zhu]{byrd1995limited}
Richard~H Byrd, Peihuang Lu, Jorge Nocedal, and Ciyou Zhu.
\newblock A limited memory algorithm for bound constrained optimization.
\newblock \emph{SIAM Journal on scientific computing}, 16\penalty0 (5):\penalty0 1190--1208, 1995.

\bibitem[Chen et~al.(2024{\natexlab{a}})Chen, Zhou, Wang, Shen, and Lyu]{chen2024trajectory}
Defang Chen, Zhenyu Zhou, Can Wang, Chunhua Shen, and Siwei Lyu.
\newblock On the trajectory regularity of ode-based diffusion sampling.
\newblock \emph{arXiv preprint arXiv:2405.11326}, 2024{\natexlab{a}}.

\bibitem[Chen et~al.(2023)Chen, Yu, Ge, Yao, Xie, Wu, Wang, Kwok, Luo, Lu, et~al.]{chen2023pixart}
Junsong Chen, Jincheng Yu, Chongjian Ge, Lewei Yao, Enze Xie, Yue Wu, Zhongdao Wang, James Kwok, Ping Luo, Huchuan Lu, et~al.
\newblock Pixart-$\alpha $: Fast training of diffusion transformer for photorealistic text-to-image synthesis.
\newblock \emph{arXiv preprint arXiv:2310.00426}, 2023.

\bibitem[Chen et~al.(2014)Chen, Fox, and Guestrin]{chen2014stochastic}
Tianqi Chen, Emily Fox, and Carlos Guestrin.
\newblock Stochastic gradient hamiltonian monte carlo.
\newblock In \emph{International conference on machine learning}, pages 1683--1691. PMLR, 2014.

\bibitem[Chen et~al.(2024{\natexlab{b}})Chen, Krishnan, Gao, Kuo, Ma, and Wang]{chen2024dsl}
Wei-Ting Chen, Gurunandan Krishnan, Qiang Gao, Sy-Yen Kuo, Sizhou Ma, and Jian Wang.
\newblock Dsl-fiqa: Assessing facial image quality via dual-set degradation learning and landmark-guided transformer.
\newblock In \emph{Proceedings of the IEEE/CVF Conference on Computer Vision and Pattern Recognition}, pages 2931--2941, 2024{\natexlab{b}}.

\bibitem[Chen et~al.(2024{\natexlab{c}})Chen, Li, Wang, Zhang, Xu, Jiang, Song, and Wang]{chen2024rethinking}
Zhichao Chen, Haoxuan Li, Fangyikang Wang, Odin Zhang, Hu Xu, Xiaoyu Jiang, Zhihuan Song, and Eric~H Wang.
\newblock Rethinking the diffusion models for numerical tabular data imputation from the perspective of wasserstein gradient flow.
\newblock \emph{arXiv preprint arXiv:2406.15762}, 2024{\natexlab{c}}.

\bibitem[Chewi et~al.(2020)Chewi, Le~Gouic, Lu, Maunu, Rigollet, and Stromme]{chewi2020exponential}
Sinho Chewi, Thibaut Le~Gouic, Chen Lu, Tyler Maunu, Philippe Rigollet, and Austin Stromme.
\newblock Exponential ergodicity of mirror-langevin diffusions.
\newblock \emph{Advances in Neural Information Processing Systems}, 33:\penalty0 19573--19585, 2020.

\bibitem[Dhariwal and Nichol(2021)]{NEURIPS2021_49ad23d1}
Prafulla Dhariwal and Alexander Nichol.
\newblock Diffusion models beat gans on image synthesis.
\newblock In \emph{Advances in Neural Information Processing Systems}, pages 8780--8794, 2021.

\bibitem[Ding(2015)]{ding2015note}
Ying Ding.
\newblock A note on quadratic transportation and divergence inequality.
\newblock \emph{Statistics \& Probability Letters}, 100:\penalty0 115--123, 2015.

\bibitem[Dockhorn et~al.(2022)Dockhorn, Vahdat, and Kreis]{dockhorn2022genie}
Tim Dockhorn, Arash Vahdat, and Karsten Kreis.
\newblock Genie: Higher-order denoising diffusion solvers.
\newblock \emph{Advances in Neural Information Processing Systems}, 35:\penalty0 30150--30166, 2022.

\bibitem[Eade(2013)]{eade2013gauss}
Ethan Eade.
\newblock Gauss-newton/levenberg-marquardt optimization.
\newblock \emph{Tech. Rep.}, 2013.

\bibitem[Fan et~al.(2019)Fan, Huang, and Pan]{fan2019adaptive}
Jinyan Fan, Jianchao Huang, and Jianyu Pan.
\newblock An adaptive multi-step levenberg--marquardt method.
\newblock \emph{Journal of Scientific Computing}, 78:\penalty0 531--548, 2019.

\bibitem[Feng et~al.(2024{\natexlab{a}})Feng, Zhao, Zhang, Dong, Ding, Jiang, and Qian]{feng2024pectp}
Qian Feng, Hanbin Zhao, Chao Zhang, Jiahua Dong, Henghui Ding, Yu-Gang Jiang, and Hui Qian.
\newblock Pectp: Parameter-efficient cross-task prompts for incremental vision transformer.
\newblock \emph{arXiv preprint arXiv:2407.03813}, 2024{\natexlab{a}}.

\bibitem[Feng et~al.(2024{\natexlab{b}})Feng, Zhou, Zhao, Zhang, and Qian]{feng2024lw2g}
Qian Feng, Dawei Zhou, Hanbin Zhao, Chao Zhang, and Hui Qian.
\newblock Lw2g: Learning whether to grow for prompt-based continual learning.
\newblock \emph{arXiv preprint arXiv:2409.18860}, 2024{\natexlab{b}}.

\bibitem[Fischer et~al.(2021)Fischer, Izmailov, and Solodov]{fischer2021unit}
Andreas Fischer, Alexey~F Izmailov, and Mikhail~V Solodov.
\newblock Unit stepsize for the newton method close to critical solutions.
\newblock \emph{Mathematical Programming}, 187\penalty0 (1):\penalty0 697--721, 2021.

\bibitem[Fu et~al.(2025)Fu, Zhao, Dong, Zhang, and Qian]{fu2025iap}
Hao Fu, Hanbin Zhao, Jiahua Dong, Chao Zhang, and Hui Qian.
\newblock Iap: Improving continual learning of vision-language models via instance-aware prompting.
\newblock \emph{arXiv preprint arXiv:2503.20612}, 2025.

\bibitem[Fu et~al.(2016)Fu, Luo, and Zhang]{fu2016quasi}
Tianfan Fu, Luo Luo, and Zhihua Zhang.
\newblock Quasi - newton hamiltonian monte carlo.
\newblock In \emph{Proceedings of the Conference on Uncertainty in Artificial Intelligence (UAI)}, 2016.

\bibitem[Gao et~al.(2023)Gao, Pan, Zhou, Kang, and Chaudhari]{gao2023fast}
Yansong Gao, Zhihong Pan, Xin Zhou, Le Kang, and Pratik Chaudhari.
\newblock Fast diffusion probabilistic model sampling through the lens of backward error analysis.
\newblock \emph{arXiv preprint arXiv:2304.11446}, 2023.

\bibitem[Gohberg et~al.(1990)Gohberg, Goldberg, Kaashoek, Gohberg, Goldberg, and Kaashoek]{gohberg1990hilbert}
Israel Gohberg, Seymour Goldberg, Marinus~A Kaashoek, Israel Gohberg, Seymour Goldberg, and Marinus~A Kaashoek.
\newblock Hilbert-schmidt operators.
\newblock \emph{Classes of Linear Operators Vol. I}, pages 138--147, 1990.

\bibitem[Hackl et~al.(1980)Hackl, Wacker, and Zulehner]{hackl1980efficient}
J Hackl, HJ Wacker, and W Zulehner.
\newblock An efficient step size control for continuation methods.
\newblock \emph{BIT Numerical Mathematics}, 20:\penalty0 475--485, 1980.

\bibitem[Hasler and Suesstrunk(2003)]{hasler2003measuring}
David Hasler and Sabine~E Suesstrunk.
\newblock Measuring colorfulness in natural images.
\newblock In \emph{Human vision and electronic imaging VIII}, pages 87--95. SPIE, 2003.

\bibitem[He et~al.(2023)He, Ming, Zheng, Zhong, and Ma]{he2023eat}
Shuai He, Anlong Ming, Shuntian Zheng, Haobin Zhong, and Huadong Ma.
\newblock Eat: An enhancer for aesthetics-oriented transformers.
\newblock In \emph{Proceedings of the 31st ACM international conference on multimedia}, pages 1023--1032, 2023.

\bibitem[Hellinger(1909)]{hellinger1909neue}
Ernst Hellinger.
\newblock Neue begr{\"u}ndung der theorie quadratischer formen von unendlichvielen ver{\"a}nderlichen.
\newblock \emph{Journal f{\"u}r die reine und angewandte Mathematik}, 1909\penalty0 (136):\penalty0 210--271, 1909.

\bibitem[Heusel et~al.(2017)Heusel, Ramsauer, Unterthiner, Nessler, and Hochreiter]{heusel2017gans}
Martin Heusel, Hubert Ramsauer, Thomas Unterthiner, Bernhard Nessler, and Sepp Hochreiter.
\newblock Gans trained by a two time-scale update rule converge to a local nash equilibrium.
\newblock \emph{Advances in Neural Information Processing Systems (NeurIPS)}, 2017.

\bibitem[Ho et~al.(2020)Ho, Jain, and Abbeel]{ho2020denoising}
Jonathan Ho, Ajay Jain, and Pieter Abbeel.
\newblock Denoising diffusion probabilistic models.
\newblock \emph{Advances in Neural Information Processing Systems}, 33:\penalty0 6840--6851, 2020.

\bibitem[Ho et~al.(2022)Ho, Saharia, Chan, Fleet, Norouzi, and Salimans]{JMLR:v23:21-0635}
Jonathan Ho, Chitwan Saharia, William Chan, David~J. Fleet, Mohammad Norouzi, and Tim Salimans.
\newblock Cascaded diffusion models for high fidelity image generation.
\newblock \emph{Journal of Machine Learning Research}, 23\penalty0 (47):\penalty0 1--33, 2022.

\bibitem[Horn and Johnson(2012)]{horn2012matrix}
Roger~A Horn and Charles~R Johnson.
\newblock \emph{Matrix analysis}.
\newblock Cambridge university press, 2012.

\bibitem[Hsieh et~al.(2018)Hsieh, Kavis, Rolland, and Cevher]{hsieh2018mirrored}
Ya-Ping Hsieh, Ali Kavis, Paul Rolland, and Volkan Cevher.
\newblock Mirrored langevin dynamics.
\newblock \emph{Advances in Neural Information Processing Systems}, 31, 2018.

\bibitem[Huang et~al.(2023)Huang, Sun, Xie, Li, and Liu]{huang2023t2i}
Kaiyi Huang, Kaiyue Sun, Enze Xie, Zhenguo Li, and Xihui Liu.
\newblock T2i-compbench: A comprehensive benchmark for open-world compositional text-to-image generation.
\newblock \emph{Advances in Neural Information Processing Systems}, 36:\penalty0 78723--78747, 2023.

\bibitem[Jolicoeur-Martineau et~al.(2021)Jolicoeur-Martineau, Li, Pich{\'e}-Taillefer, Kachman, and Mitliagkas]{jolicoeur2021gotta}
Alexia Jolicoeur-Martineau, Ke Li, R{\'e}mi Pich{\'e}-Taillefer, Tal Kachman, and Ioannis Mitliagkas.
\newblock Gotta go fast when generating data with score-based models.
\newblock \emph{arXiv preprint arXiv:2105.14080}, 2021.

\bibitem[Jordan et~al.(1998)Jordan, Kinderlehrer, and Otto]{jordan1998variational}
Richard Jordan, David Kinderlehrer, and Felix Otto.
\newblock The variational formulation of the fokker--planck equation.
\newblock \emph{SIAM journal on mathematical analysis}, 29\penalty0 (1):\penalty0 1--17, 1998.

\bibitem[Kac(1987)]{kac1987enigmas}
Mark Kac.
\newblock \emph{Enigmas of chance: an autobiography}.
\newblock Univ of California Press, 1987.

\bibitem[Karatzas and Shreve(2014)]{karatzas2014brownian}
Ioannis Karatzas and Steven Shreve.
\newblock \emph{Brownian motion and stochastic calculus}.
\newblock springer, 2014.

\bibitem[Karras et~al.(2017)Karras, Aila, Laine, and Lehtinen]{karras2017progressive}
Tero Karras, Timo Aila, Samuli Laine, and Jaakko Lehtinen.
\newblock Progressive growing of gans for improved quality, stability, and variation.
\newblock \emph{arXiv preprint arXiv:1710.10196}, 2017.

\bibitem[Karras et~al.(2022)Karras, Aittala, Aila, and Laine]{karras2022elucidating}
Tero Karras, Miika Aittala, Timo Aila, and Samuli Laine.
\newblock Elucidating the design space of diffusion-based generative models.
\newblock \emph{arXiv preprint arXiv:2206.00364}, 2022.

\bibitem[Kawamoto(2009)]{kawamoto2009stabilization}
Atsushi Kawamoto.
\newblock Stabilization of geometrically nonlinear topology optimization by the levenberg--marquardt method.
\newblock \emph{Structural and Multidisciplinary Optimization}, 37:\penalty0 429--433, 2009.

\bibitem[Kirstain et~al.(2023)Kirstain, Polyak, Singer, Matiana, Penna, and Levy]{kirstain2023pick}
Yuval Kirstain, Adam Polyak, Uriel Singer, Shahbuland Matiana, Joe Penna, and Omer Levy.
\newblock Pick-a-pic: An open dataset of user preferences for text-to-image generation.
\newblock \emph{Advances in Neural Information Processing Systems}, 36:\penalty0 36652--36663, 2023.

\bibitem[Ku and Yeih(2012)]{ku2012dynamical}
Cheng-Yu Ku and Weichung Yeih.
\newblock Dynamical newton-like methods with adaptive stepsize for solving nonlinear algebraic equations.
\newblock \emph{Computers, Materials, \& Continua}, 31\penalty0 (3):\penalty0 173--200, 2012.

\bibitem[Kullback and Leibler(1951)]{kullback1951information}
Solomon Kullback and Richard~A Leibler.
\newblock On information and sufficiency.
\newblock \emph{The annals of mathematical statistics}, 22\penalty0 (1):\penalty0 79--86, 1951.

\bibitem[Levenberg(1944)]{levenberg1944method}
Kenneth Levenberg.
\newblock A method for the solution of certain non-linear problems in least squares.
\newblock \emph{Quarterly of applied mathematics}, 2\penalty0 (2):\penalty0 164--168, 1944.

\bibitem[Liao et~al.(2024)Liao, Xie, Chen, Lu, and Deng]{liao2024facescore}
Zhenyi Liao, Qingsong Xie, Chen Chen, Hannan Lu, and Zhijie Deng.
\newblock Facescore: Benchmarking and enhancing face quality in human generation.
\newblock \emph{arXiv preprint arXiv:2406.17100}, 2024.

\bibitem[Lin et~al.(2014)Lin, Maire, Belongie, Hays, Perona, Ramanan, Doll{\'a}r, and Zitnick]{lin2014microsoft}
Tsung-Yi Lin, Michael Maire, Serge Belongie, James Hays, Pietro Perona, Deva Ramanan, Piotr Doll{\'a}r, and C~Lawrence Zitnick.
\newblock Microsoft coco: Common objects in context.
\newblock In \emph{Computer Vision--ECCV 2014: 13th European Conference, Zurich, Switzerland, September 6-12, 2014, Proceedings, Part V 13}, pages 740--755. Springer, 2014.

\bibitem[Lipman et~al.(2022)Lipman, Chen, Ben-Hamu, Nickel, and Le]{lipman2022flow}
Yaron Lipman, Ricky~TQ Chen, Heli Ben-Hamu, Maximilian Nickel, and Matt Le.
\newblock Flow matching for generative modeling.
\newblock \emph{arXiv preprint arXiv:2210.02747}, 2022.

\bibitem[Liu et~al.(2023)Liu, Chen, Theodorou, and Tao]{liu2023mirror}
Guan-Horng Liu, Tianrong Chen, Evangelos Theodorou, and Molei Tao.
\newblock Mirror diffusion models for constrained and watermarked generation.
\newblock \emph{Advances in Neural Information Processing Systems}, 36:\penalty0 42898--42917, 2023.

\bibitem[Liu et~al.(2022)Liu, Ren, Lin, and Zhao]{liu2022pseudo}
Luping Liu, Yi Ren, Zhijie Lin, and Zhou Zhao.
\newblock Pseudo numerical methods for diffusion models on manifolds.
\newblock \emph{arXiv preprint arXiv:2202.09778}, 2022.

\bibitem[Lu et~al.(2022{\natexlab{a}})Lu, Zhou, Bao, Chen, Li, and Zhu]{lu2022dpm}
Cheng Lu, Yuhao Zhou, Fan Bao, Jianfei Chen, Chongxuan Li, and Jun Zhu.
\newblock Dpm-solver: A fast ode solver for diffusion probabilistic model sampling in around 10 steps.
\newblock \emph{Advances in Neural Information Processing Systems}, 35:\penalty0 5775--5787, 2022{\natexlab{a}}.

\bibitem[Lu et~al.(2022{\natexlab{b}})Lu, Zhou, Bao, Chen, Li, and Zhu]{lu2022dpm++}
Cheng Lu, Yuhao Zhou, Fan Bao, Jianfei Chen, Chongxuan Li, and Jun Zhu.
\newblock Dpm-solver++: Fast solver for guided sampling of diffusion probabilistic models.
\newblock \emph{arXiv preprint arXiv:2211.01095}, 2022{\natexlab{b}}.

\bibitem[Marquardt(1963)]{marquardt1963algorithm}
Donald~W Marquardt.
\newblock An algorithm for least-squares estimation of nonlinear parameters.
\newblock \emph{Journal of the society for Industrial and Applied Mathematics}, 11\penalty0 (2):\penalty0 431--441, 1963.

\bibitem[Martin et~al.(2012)Martin, Wilcox, Burstedde, and Ghattas]{martin2012stochastic}
James Martin, Lucas~C Wilcox, Carsten Burstedde, and Omar Ghattas.
\newblock A stochastic newton mcmc method for large-scale statistical inverse problems with application to seismic inversion.
\newblock \emph{SIAM Journal on Scientific Computing}, 34\penalty0 (3):\penalty0 A1460--A1487, 2012.

\bibitem[Nemirovskij and Yudin(1983)]{nemirovskij1983problem}
Arkadij~Semenovi{\v{c}} Nemirovskij and David~Borisovich Yudin.
\newblock Problem complexity and method efficiency in optimization.
\newblock 1983.

\bibitem[Nesterov et~al.(2018)]{nesterov2018lectures}
Yurii Nesterov et~al.
\newblock \emph{Lectures on convex optimization}.
\newblock Springer, 2018.

\bibitem[Ngia and Sjoberg(2000)]{ngia2000efficient}
Lester~SH Ngia and Jonas Sjoberg.
\newblock Efficient training of neural nets for nonlinear adaptive filtering using a recursive levenberg-marquardt algorithm.
\newblock \emph{IEEE Transactions on Signal Processing}, 48\penalty0 (7):\penalty0 1915--1927, 2000.

\bibitem[Nichol et~al.(2022)Nichol, Dhariwal, Ramesh, Shyam, Mishkin, Mcgrew, Sutskever, and Chen]{pmlr-v162-nichol22a}
Alexander~Quinn Nichol, Prafulla Dhariwal, Aditya Ramesh, Pranav Shyam, Pamela Mishkin, Bob Mcgrew, Ilya Sutskever, and Mark Chen.
\newblock {GLIDE}: Towards photorealistic image generation and editing with text-guided diffusion models.
\newblock In \emph{Proceedings of the 39th International Conference on Machine Learning}, pages 16784--16804, 2022.

\bibitem[Podell et~al.(2023)Podell, English, Lacey, Blattmann, Dockhorn, M{\"u}ller, Penna, and Rombach]{podell2023sdxl}
Dustin Podell, Zion English, Kyle Lacey, Andreas Blattmann, Tim Dockhorn, Jonas M{\"u}ller, Joe Penna, and Robin Rombach.
\newblock Sdxl: Improving latent diffusion models for high-resolution image synthesis.
\newblock \emph{arXiv preprint arXiv:2307.01952}, 2023.

\bibitem[Polyak(2007)]{polyak2007newton}
Boris~T Polyak.
\newblock Newton’s method and its use in optimization.
\newblock \emph{European Journal of Operational Research}, 181\penalty0 (3):\penalty0 1086--1096, 2007.

\bibitem[Radford et~al.(2021)Radford, Kim, Hallacy, Ramesh, Goh, Agarwal, Sastry, Askell, Mishkin, Clark, et~al.]{radford2021learning}
Alec Radford, Jong~Wook Kim, Chris Hallacy, Aditya Ramesh, Gabriel Goh, Sandhini Agarwal, Girish Sastry, Amanda Askell, Pamela Mishkin, Jack Clark, et~al.
\newblock Learning transferable visual models from natural language supervision.
\newblock In \emph{International Conference on Machine Learning (ICML)}, pages 8748--8763, 2021.

\bibitem[Ramesh et~al.(2022)Ramesh, Dhariwal, Nichol, Chu, and Chen]{ramesh2022hierarchical}
Aditya Ramesh, Prafulla Dhariwal, Alex Nichol, Casey Chu, and Mark Chen.
\newblock Hierarchical text-conditional image generation with clip latents.
\newblock \emph{arXiv preprint arXiv:2204.06125}, 2022.

\bibitem[Risken(1996)]{risken1996fokker}
H Risken.
\newblock The fokker-planck equation, 1996.

\bibitem[Rissanen et~al.(2025)Rissanen, Heinonen, and Solin]{rissanen2025free}
Severi Rissanen, Markus Heinonen, and Arno Solin.
\newblock Free hunch: Denoiser covariance estimation for diffusion models without extra costs.
\newblock In \emph{The Thirteenth International Conference on Learning Representations}, 2025.

\bibitem[Robert(1999)]{robert1999monte}
CP Robert.
\newblock Monte carlo statistical methods, 1999.

\bibitem[Rockafellar(2015)]{rockafellar2015convex}
Ralph~Tyrell Rockafellar.
\newblock Convex analysis:(pms-28).
\newblock 2015.

\bibitem[Rombach et~al.(2022)Rombach, Blattmann, Lorenz, Esser, and Ommer]{rombach2022high}
Robin Rombach, Andreas Blattmann, Dominik Lorenz, Patrick Esser, and Bj{\"o}rn Ommer.
\newblock High-resolution image synthesis with latent diffusion models.
\newblock In \emph{Proceedings of the IEEE/CVF Conference on Computer Vision and Pattern Recognition}, pages 10684--10695, 2022.

\bibitem[Roweis(1996)]{roweis1996levenberg}
Sam Roweis.
\newblock Levenberg-marquardt optimization.
\newblock \emph{Notes, University Of Toronto}, 52, 1996.

\bibitem[Saharia et~al.(2022{\natexlab{a}})Saharia, Chan, Saxena, Li, Whang, Denton, Ghasemipour, Gontijo-Lopes, Ayan, Salimans, Ho, Fleet, and Norouzi]{saharia2022photorealistic}
Chitwan Saharia, William Chan, Saurabh Saxena, Lala Li, Jay Whang, Emily Denton, Seyed Kamyar~Seyed Ghasemipour, Raphael Gontijo-Lopes, Burcu~Karagol Ayan, Tim Salimans, Jonathan Ho, David~J. Fleet, and Mohammad Norouzi.
\newblock Photorealistic text-to-image diffusion models with deep language understanding.
\newblock In \emph{Advances in Neural Information Processing Systems (NeurIPS)}, pages 36479--36494, 2022{\natexlab{a}}.

\bibitem[Saharia et~al.(2022{\natexlab{b}})Saharia, Chan, Saxena, Li, Whang, Denton, Ghasemipour, Gontijo~Lopes, Karagol~Ayan, Salimans, Ho, Fleet, and Norouzi]{NEURIPS2022_ec795aea}
Chitwan Saharia, William Chan, Saurabh Saxena, Lala Li, Jay Whang, Emily~L Denton, Kamyar Ghasemipour, Raphael Gontijo~Lopes, Burcu Karagol~Ayan, Tim Salimans, Jonathan Ho, David~J Fleet, and Mohammad Norouzi.
\newblock Photorealistic text-to-image diffusion models with deep language understanding.
\newblock In \emph{Advances in Neural Information Processing Systems}, pages 36479--36494, 2022{\natexlab{b}}.

\bibitem[Schuhmann et~al.(2022)Schuhmann, Beaumont, Vencu, Gordon, Wightman, Cherti, Coombes, Katta, Mullis, Wortsman, et~al.]{schuhmann2022laion}
Christoph Schuhmann, Romain Beaumont, Richard Vencu, Cade Gordon, Ross Wightman, Mehdi Cherti, Theo Coombes, Aarush Katta, Clayton Mullis, Mitchell Wortsman, et~al.
\newblock Laion-5b: An open large-scale dataset for training next generation image-text models.
\newblock \emph{Advances in neural information processing systems}, 35:\penalty0 25278--25294, 2022.

\bibitem[Sherman and Morrison(1950)]{sherman1950adjustment}
Jack Sherman and Winifred~J Morrison.
\newblock Adjustment of an inverse matrix corresponding to a change in one element of a given matrix.
\newblock \emph{The Annals of Mathematical Statistics}, 21\penalty0 (1):\penalty0 124--127, 1950.

\bibitem[Simsekli et~al.(2016)Simsekli, Badeau, Cemgil, and Richard]{simsekli2016stochastic}
Umut Simsekli, Roland Badeau, Taylan Cemgil, and Ga{\"e}l Richard.
\newblock Stochastic quasi-newton langevin monte carlo.
\newblock In \emph{International Conference on Machine Learning}, pages 642--651. PMLR, 2016.

\bibitem[Sohl-Dickstein et~al.(2015)Sohl-Dickstein, Weiss, Maheswaranathan, and Ganguli]{sohl2015deep}
Jascha Sohl-Dickstein, Eric Weiss, Niru Maheswaranathan, and Surya Ganguli.
\newblock Deep unsupervised learning using nonequilibrium thermodynamics.
\newblock In \emph{International Conference on Machine Learning}, pages 2256--2265. PMLR, 2015.

\bibitem[Song et~al.(2021)Song, Meng, and Ermon]{song2021denoising}
Jiaming Song, Chenlin Meng, and Stefano Ermon.
\newblock Denoising diffusion implicit models.
\newblock In \emph{International Conference on Learning Representations}, 2021.

\bibitem[Song and Ermon(2019)]{song2019generative}
Yang Song and Stefano Ermon.
\newblock Generative modeling by estimating gradients of the data distribution.
\newblock \emph{Advances in neural information processing systems}, 32, 2019.

\bibitem[Song et~al.(2020)Song, Sohl-Dickstein, Kingma, Kumar, Ermon, and Poole]{song2020score}
Yang Song, Jascha Sohl-Dickstein, Diederik~P Kingma, Abhishek Kumar, Stefano Ermon, and Ben Poole.
\newblock Score-based generative modeling through stochastic differential equations.
\newblock \emph{arXiv preprint arXiv:2011.13456}, 2020.

\bibitem[Tu et~al.(2025{\natexlab{a}})Tu, Fu, Yang, Zhao, Zhang, and Qian]{tu2025texttoucher}
Jiahang Tu, Hao Fu, Fengyu Yang, Hanbin Zhao, Chao Zhang, and Hui Qian.
\newblock Texttoucher: Fine-grained text-to-touch generation.
\newblock In \emph{Proceedings of the AAAI Conference on Artificial Intelligence}, pages 7455--7463, 2025{\natexlab{a}}.

\bibitem[Tu et~al.(2025{\natexlab{b}})Tu, Ji, Zhao, Zhang, Zimmermann, and Qian]{tu2025driveditfit}
Jiahang Tu, Wei Ji, Hanbin Zhao, Chao Zhang, Roger Zimmermann, and Hui Qian.
\newblock Driveditfit: Fine-tuning diffusion transformers for autonomous driving data generation.
\newblock \emph{ACM Transactions on Multimedia Computing, Communications and Applications}, 21\penalty0 (3):\penalty0 1--29, 2025{\natexlab{b}}.

\bibitem[Verd{\'u}(2014)]{verdu2014total}
Sergio Verd{\'u}.
\newblock Total variation distance and the distribution of relative information.
\newblock In \emph{2014 information theory and applications workshop (ITA)}, pages 1--3. IEEE, 2014.

\bibitem[Villani et~al.(2009)]{villani2009optimal}
C{\'e}dric Villani et~al.
\newblock \emph{Optimal transport: old and new}.
\newblock Springer, 2009.

\bibitem[Wang et~al.(2024{\natexlab{a}})Wang, Yin, Dong, Zhu, Zhang, Zhao, Qian, and Li]{wang2024belm}
Fangyikang Wang, Hubery Yin, Yuejiang Dong, Huminhao Zhu, Chao Zhang, Hanbin Zhao, Hui Qian, and Chen Li.
\newblock Belm: Bidirectional explicit linear multi-step sampler for exact inversion in diffusion models.
\newblock \emph{arXiv preprint arXiv:2410.07273}, 2024{\natexlab{a}}.

\bibitem[Wang et~al.(2024{\natexlab{b}})Wang, Zhu, Zhang, Zhao, and Qian]{wang2024gad}
Fangyikang Wang, Huminhao Zhu, Chao Zhang, Hanbin Zhao, and Hui Qian.
\newblock Gad-pvi: A general accelerated dynamic-weight particle-based variational inference framework.
\newblock \emph{Entropy}, 26\penalty0 (8):\penalty0 679, 2024{\natexlab{b}}.

\bibitem[Wang et~al.(2025)Wang, Yin, Zhuang, Zhu, Li, Qian, Zhang, Zhao, Qian, and Li]{wang2025efficientlyaccessdiffusionfisher}
Fangyikang Wang, Hubery Yin, Shaobin Zhuang, Huminhao Zhu, Yinan Li, Lei Qian, Chao Zhang, Hanbin Zhao, Hui Qian, and Chen Li.
\newblock Efficiently access diffusion fisher: Within the outer product span space, 2025.

\bibitem[Welling and Teh(2011)]{welling2011bayesian}
Max Welling and Yee~W Teh.
\newblock Bayesian learning via stochastic gradient langevin dynamics.
\newblock In \emph{Proceedings of the 28th international conference on machine learning (ICML-11)}, pages 681--688. Citeseer, 2011.

\bibitem[Xia et~al.(2024)Xia, Shen, Lei, Zhou, Zhao, Yi, Wang, and Liu]{xia2024towards}
Mengfei Xia, Yujun Shen, Changsong Lei, Yu Zhou, Deli Zhao, Ran Yi, Wenping Wang, and Yong-Jin Liu.
\newblock Towards more accurate diffusion model acceleration with a timestep tuner.
\newblock In \emph{Proceedings of the IEEE/CVF Conference on Computer Vision and Pattern Recognition}, pages 5736--5745, 2024.

\bibitem[Xue et~al.(2024)Xue, Liu, Chen, Zhang, Hu, Xie, and Li]{xue2024accelerating}
Shuchen Xue, Zhaoqiang Liu, Fei Chen, Shifeng Zhang, Tianyang Hu, Enze Xie, and Zhenguo Li.
\newblock Accelerating diffusion sampling with optimized time steps.
\newblock In \emph{Proceedings of the IEEE/CVF Conference on Computer Vision and Pattern Recognition}, pages 8292--8301, 2024.

\bibitem[Zhang et~al.(2023)Zhang, Rao, and Agrawala]{zhang2023adding}
Lvmin Zhang, Anyi Rao, and Maneesh Agrawala.
\newblock Adding conditional control to text-to-image diffusion models.
\newblock In \emph{Proceedings of the IEEE/CVF International Conference on Computer Vision}, pages 3836--3847, 2023.

\bibitem[Zhang et~al.(2024)Zhang, Yin, Li, and Xie]{zhang2024tackling}
Pengze Zhang, Hubery Yin, Chen Li, and Xiaohua Xie.
\newblock Tackling the singularities at the endpoints of time intervals in diffusion models.
\newblock \emph{arXiv preprint arXiv:2403.08381}, 2024.

\bibitem[Zhang and Chen(2022)]{zhang2022fast}
Qinsheng Zhang and Yongxin Chen.
\newblock Fast sampling of diffusion models with exponential integrator.
\newblock \emph{arXiv preprint arXiv:2204.13902}, 2022.

\bibitem[Zhao et~al.(2024)Zhao, Bai, Rao, Zhou, and Lu]{zhao2024unipc}
Wenliang Zhao, Lujia Bai, Yongming Rao, Jie Zhou, and Jiwen Lu.
\newblock Unipc: A unified predictor-corrector framework for fast sampling of diffusion models.
\newblock \emph{Advances in Neural Information Processing Systems}, 36, 2024.

\bibitem[Zheng et~al.(2023)Zheng, Lu, Chen, and Zhu]{zheng2023dpm}
Kaiwen Zheng, Cheng Lu, Jianfei Chen, and Jun Zhu.
\newblock Dpm-solver-v3: Improved diffusion ode solver with empirical model statistics.
\newblock \emph{Advances in Neural Information Processing Systems}, 36:\penalty0 55502--55542, 2023.

\bibitem[Zhou et~al.(2024)Zhou, Chen, Wang, and Chen]{zhou2024fast}
Zhenyu Zhou, Defang Chen, Can Wang, and Chun Chen.
\newblock Fast ode-based sampling for diffusion models in around 5 steps.
\newblock In \emph{Proceedings of the IEEE/CVF Conference on Computer Vision and Pattern Recognition}, pages 7777--7786, 2024.

\bibitem[Zhu et~al.(2024{\natexlab{a}})Zhu, Wang, Ding, Qu, and Zhu]{zhu2024analyzing}
Huminhao Zhu, Fangyikang Wang, Tianyu Ding, Qing Qu, and Zhihui Zhu.
\newblock Analyzing and improving model collapse in rectified flow models.
\newblock \emph{arXiv preprint arXiv:2412.08175}, 2024{\natexlab{a}}.

\bibitem[Zhu et~al.(2024{\natexlab{b}})Zhu, Wang, Zhang, Zhao, and Qian]{zhu2024neural}
Huminhao Zhu, Fangyikang Wang, Chao Zhang, Hanbin Zhao, and Hui Qian.
\newblock Neural sinkhorn gradient flow.
\newblock \emph{arXiv preprint arXiv:2401.14069}, 2024{\natexlab{b}}.

\end{thebibliography}
}

\onecolumn
\clearpage
\setcounter{page}{1}
\maketitlesupplementary
\appendix

\section{Proofs and Detailed Formulations}
\subsection{Formulations of gradient descent and Newton's method}
Gradient Descent (GD) \citep{nesterov2018lectures} is a renowned optimization method for finding the minimum.
Given a differentiable function $f$, its iterative scheme writes
\begin{equation} 
  \vx_{k+1} = \vx_{k} - \eta\nabla_\vx f (\vx_k).
\end{equation}

Given a twice differentiable function $f$, the iterative scheme of Newton's method writes
\begin{equation} 
  \vx_{k+1} = \vx_{k} - \eta\left[\mH_f(\vx_k) \right]^{-1}\nabla_\vx f (\vx_k).
\end{equation}
In statistics, the \textit{Hessian geometry} of a distribution $p(\vx)$ is defined to be $\nabla^2_\vx \log p(\vx)$. 

\subsection{Proof of Proposition \ref{prop:gn_appro} }\label{supp:gn_proof}
Here, we give the proof of Proposition \ref{prop:gn_appro}, which is an analogy of the Gauss-Newton technique in the diffusion model context.
\begin{proof}

\begin{equation} \label{gn_1}
 p_t\left(\vx_t\right) \approx \frac{1}{\sqrt{2p}\sigma_t}\exp \left(-\frac{\norm{\vx_t-\alpha_t \vy_\theta(\vx_t,t)}^2}{2\sigma_t^2}\right),
\end{equation} 
denote $r(\vx_t)=\norm{\vx_t-\alpha_t \vy_\theta(\vx_t,t)}\in \sR$

\begin{equation} \label{gn_2}
\begin{aligned}
 -\frac{\vepsilon_\theta}{\sigma_t}&=\nabla_{\vx_t}\log p_t\left(\vx_t\right)\\
 & = -\frac{1}{2\sigma_t^2} \nabla_{\vx_t}\norm{\vx_t-\alpha_t \vy_\theta(\vx_t,t)}^2 \\
 & = -\frac{1}{\sigma_t^2}r(\vx_t)\frac{\partial r(\vx_t)}{\partial \vx_t}.
\end{aligned}
\end{equation} 
Rearranging Eq. \ref{gn_2}, we can use $\vepsilon_\theta$ to represent $\frac{\partial r(\vx_t)}{\partial \vx_t}$:
\begin{equation} \label{gn_3}
\begin{aligned}
    \frac{\partial r(\vx_t)}{\partial \vx_t} = \frac{\sigma_t}{r(\vx_t)}\vepsilon_\theta.
\end{aligned}
\end{equation} 
We can then approximate the Jacobian matrix of $\vepsilon_\theta$ using the Gauss-Newton technique, which essence in the omission of the second derivatives $\frac{\partial^2 r(\vx_t)}{\partial \vx_t^2}$.
\begin{equation} \label{gn_4}
\begin{aligned}
    \frac{\partial \vepsilon_\theta(\vx_t,t)}{\partial \vx_t} &=\frac{\partial \left(\frac{r(\vx_t)}{\sigma_t}\frac{\partial r(\vx_t)}{\partial \vx_t}\right)}{\partial \vx_t}\\
    &=\frac{1}{\sigma_t}\left[\frac{\partial r(\vx_t)}{\partial \vx_t}\frac{\partial r(\vx_t)}{\partial \vx_t}^T+r(\vx_t)\frac{\partial^2 r(\vx_t)}{\partial \vx_t^2} \right]\\
    &\approx \frac{1}{\sigma_t}\frac{\partial r(\vx_t)}{\partial \vx_t}\frac{\partial r(\vx_t)}{\partial \vx_t}^T\\
    &=\frac{1}{\sigma_t}\left( \frac{\sigma_t}{r(\vx_t)}\vepsilon_\theta\right)\left( \frac{\sigma_t}{r(\vx_t)}\vepsilon_\theta\right)^T\\
    &=\frac{\sigma_t}{\norm{\vx_t-\alpha_t \vy_\theta(\vx_t,t)}^2}\vepsilon_\theta\vepsilon_\theta^T\\
    &=\frac{\sigma_t}{\norm{\sigma_t\vepsilon_\theta}^2}\vepsilon_\theta\vepsilon_\theta^T\\
    &=\frac{1}{\sigma_t\norm{\vepsilon_\theta}^2}\vepsilon_\theta\vepsilon_\theta^T.
\end{aligned}
\end{equation} 

\end{proof}

\subsection{Detailed Derivation of Eq. \ref{sm-formula}}
The Eq. \ref{sm-formula} is a direct result of the Sherman–Morrison formula. Here we first state this formula and then derive Eq. \ref{sm-formula}.
\begin{theorem} 
(Sherman–Morrison formula, \cite{sherman1950adjustment})
Suppose $A \in \mathbb{R}^{n \times n}$ is an invertible square matrix and $u, v \in \mathbb{R}^n$ are column vectors. Then $A+u v^{\top}$ is invertible iff $1+v^{\top} A^{-1} u \neq 0$. In this case,
\begin{equation} \label{morrison}
\begin{aligned}
\left(A+u v^{\top}\right)^{-1}=A^{-1}-\frac{A^{-1} u v^{\top} A^{-1}}{1+v^{\top} A^{-1} u}.
\end{aligned}
\end{equation} 
Here, $u v^{\top}$ is the outer product of two vectors $u$ and $v$.
\end{theorem}

The Eq. \ref{sm-formula} can be obtained by applying the Sherman–Morrison formula with $A = \lambda \mI$ and $u=v=\vepsilon_\theta$. In this case, the requirements of the Sherman–Morrison formula are satisfied, as $1+\vepsilon_\theta^{\top} \left(\lambda \mI\right)^{-1} \vepsilon_\theta = 1+\frac{\norm{\vepsilon_\theta}^2}{\lambda} \neq 0$.

\subsection{Derivation of Eq. \ref{lm_reverse_sde} }
Our way of transforming the annealing Langevin to the continuous SDE adopts a technique akin to that in \cite{song2020score}.
Discretizing the Eq. \ref{lml_continuous} at each nose level, and gradually decreasing the noise level, we would get the following iterative schemes:
\begin{equation} \label{anneal_discrete}
\vx_i= \vx_{i-1} + \epsilon_i \mH_{LM}^{-1}(\vx_{i+1},\lambda)\nabla_{\vx} \log p\left(\vx_{i+1}\right)  +\sqrt{2\epsilon_i}\mH_{LM}^{-1}\vz_i,
\end{equation} 
Then, applying the reverse process of the ancestral sampling \cite{song2019generative} to Eq. \ref{anneal_discrete}, we obtain the LM annealing SDE in Eq. \ref{lm_reverse_sde}.

\subsection{Derivation of Eq. \ref{lm_reverse_ode} }
Here, we will prove that the marginal distribution of Eq. \ref{lm_reverse_ode} is the same as that of Eq. \ref{lm_reverse_sde}.
We will first state the Feynman–Kac formula.
\begin{theorem} \label{them:kac}
(Feynman–Kac formula, \cite{kac1987enigmas})
For an Ito SDE as follows 
\begin{equation} 
\begin{aligned}
\rd \mathbf{x}_t=\boldsymbol{\mu}\left(\mathbf{x}_t, t\right) \rd t+\boldsymbol{\sigma}\left(\mathbf{x}_t, t\right) \rd \mathbf{B}_t,
\end{aligned}
\end{equation} 
Its underlying distribution evolves according to the following Fokker-Planck equation:
\begin{equation} 
\begin{aligned}
\frac{\partial p(\mathbf{x}, t)}{\partial t}=-\nabla \cdot[\boldsymbol{\mu} p]+\frac{1}{2} \nabla \cdot\left(\boldsymbol{\sigma} \boldsymbol{\sigma}^{\top} \nabla p\right)
\end{aligned}
\end{equation} 
\end{theorem}
Applying the theorem \ref{them:kac}, the Fokker-Planck equation of Eq. \ref{lm_reverse_sde} writes:
\begin{equation} 
\small
\begin{aligned}
&\frac{\partial p_t}{\partial t}\\
=&-\nabla \cdot[f_t \vx_t p_t-g^2_t\mH_{LM}^{-1}\nabla_{\vx_t}\log p_t \cdot p_t]+\frac{1}{2} \nabla \cdot\left(g^2_t\mH_{LM}^{-1} \nabla p_t\right)\\
=&-\nabla \cdot[f_t \vx_t p_t-g^2_t\mH_{LM}^{-1}\nabla p_t]+\frac{1}{2} \nabla \cdot\left(g^2_t \mH_{LM}^{-1}\nabla p_t\right)\\
=&-\nabla \cdot[f_t \vx_t p_t-\frac{1}{2}g^2_t\mH_{LM}^{-1}\nabla p_t]\\
\end{aligned}
\end{equation} 
Applying the theorem \ref{them:kac}, the Fokker-Planck equation of Eq. \ref{lm_reverse_ode} writes:
\begin{equation} 
\begin{aligned}
\frac{\partial p_t}{\partial t}
=-\nabla \cdot[f_t \vx_t p_t-\frac{1}{2}g^2_t\mH_{LM}^{-1}\nabla p_t]\\
\end{aligned}
\end{equation} 

It is obvious to see that Eq. \ref{lm_reverse_sde} and Eq. \ref{lm_reverse_ode} result in the same form of Fokker-Planck equation. As they start from the same initial noise distribution, their marginal distribution will also be the same.

\subsection{Proof of Proposition \ref{prop:error_bound_hessian} }\label{supp:error_proof}
\begin{proof}
   In order to establish the boundary for the Gauss-Newton-type technique in diffusion scenarios, it is necessary to estimate the boundary of the simplified second derivatives. These second derivatives incorporate the Fisher information of the diffused distribution. We therefore utilize the analytical form as outlined in Proposition 3 of \cite{wang2025efficientlyaccessdiffusionfisher}, leveraging their form to establish the boundary. The notation used in this proof is borrowed from their work.
\begin{equation} 
\begin{aligned}
&\norm{r(\vx_t)\frac{\partial^2 r(\vx_t)}{\partial \vx_t^2}}_{HS}\\
  \leq& \norm{r(\vx_t)}\norm{\frac{\partial^2 r(\vx_t)}{\partial \vx_t^2}}_{HS} \\
  \leq& \norm{\vx_t-\alpha_t \vy_\theta(\vx_t,t)} \left(\delta_3 + 2\sigma_t\norm{F_t(\vx_t)}_{HS}\right)\\
  \leq& \left(\alpha_t\delta_2+\norm{\vx_t-\alpha_t\bar{\vy}}\right) \\
  &\left(\delta_3 + 2\sigma_t^2\norm{\frac{1}{\sigma_t^2}\mI}_{HS}+ 2\sigma_t^2\norm{\frac{\alpha_t^2}{\sigma_t^4}\sum_i w_i \vy_i\vy_i^{\top}}_{HS}\right.
  \\&\left.+2\sigma_t^2\norm{\frac{\alpha_t^2}{\sigma_t^4}\left(\sum_i w_i \vy_i\right)\left(\sum_i w_i \vy_i\right)^{\top}}_{HS}\right)\\
  \leq & \left(\alpha_t\delta_2+\delta_1+\alpha_t\mathcal{D}_y \right)\left( \delta_3 + 2+2\frac{\alpha_t^2}{\sigma_t^2} \mathcal{D}_y^2\right)
\end{aligned}
\end{equation}  
\end{proof}

\subsection{Proof of Proposition \ref{prop:stationary} }\label{supp:stationary_proof}
To obtain the stationary analysis result of our LM-Langevin, we first borrow the stationary analysis of mirror Langevin from \cite{hsieh2018mirrored}.
\begin{theorem}
    \cite[Eq 3.2]{hsieh2018mirrored}
    If we are able to draw a sample $\mathbf{Y}$ from $e^{-W(\mathbf{y})} \mathrm{d} \mathbf{y}$, then $\nabla h^{\star}(\mathbf{Y})$ immediately gives a sample for the desired distribution $e^{-V(\mathbf{x})} \mathrm{d} \mathbf{x}$. Furthermore, suppose for the moment that $\operatorname{dom}\left(h^{\star}\right)=\mathbb{R}^d$, so that $e^{-W(\mathbf{y})} \mathrm{d} \mathbf{y}$ is unconstrained. Then we can simply exploit the classical Langevin Dynamics (1.1) to efficiently take samples from $e^{-W(\mathbf{y})} \mathrm{d} \mathbf{y}$.
The above reasoning leads us to set up the Mirrored Langevin Dynamics (MLD):
\begin{equation} 
\left\{\begin{array}{l}
\mathrm{d} \mathbf{Y}_t=-(\nabla W \circ \nabla h)\left(\mathbf{X}_t\right) \mathrm{d} t+\sqrt{2} \mathrm{~d} \mathbf{B}_t \\
\mathbf{X}_t=\nabla h^{\star}\left(\mathbf{Y}_t\right)
\end{array}\right.
\end{equation} 
Notice that the stationary distribution of $\mathbf{Y}_t$ in MLD is $e^{-W(\mathbf{y})} \mathrm{d} \mathbf{y}$.
\end{theorem}
Then the behavior of our stationary in Proposition \ref{prop:stationary} is a direct result of Theorem 2 by setting the mirror duality map $\nabla h$ as $\nabla \log p_t(\cdot)+\lambda\norm{\cdot}$.

\subsection{Proof of Proposition \ref{prop:lmd_convergence} }\label{supp:convergence_proof}

\begin{definition}
The $\chi^2$-distance between $\mu$ and $\pi$ is defined as :
\begin{equation} \label{chi}
\begin{aligned}
\chi^2(\mu \| \pi):=\operatorname{var}_\pi \frac{\mathrm{d} \mu}{\mathrm{d} \pi}=\int\left(\frac{\mathrm{d} \mu}{\mathrm{d} \pi}\right)^2 \mathrm{~d} \pi-1, \quad \text{if} \quad \mu \ll \pi
\end{aligned}
\end{equation} 
and $\chi^2(\mu \| \pi) = \infty$ otherwise, where $ \mu \ll \pi$ means $\mu$ is absolutely continuous with respect to $\pi$.
\end{definition}

\begin{definition}
(Mirror Poincaré condition, \cite{chewi2020exponential}). Given a mirror map $\phi$,  that is a strictly convex twice continuously differentiable function of Legendre type \citep{rockafellar2015convex}, we say that the distribution $\pi$ satisfies a mirror Poincaré condition with constant $C_{\mathrm{MP}}$ if
\begin{equation} \label{}
\begin{aligned}
(\text{MP})~~~~~~    \operatorname{var}_\pi g \leq C_{\mathrm{MP}} \mathbb{E}_\pi\left\langle\nabla g,\left(\nabla^2 \phi\right)^{-1} \nabla g\right\rangle,
\\ \text { for all locally Lipschitz } g \in L^2(\pi)
\end{aligned}
\end{equation} 
When $\phi=\|\cdot\|^2 / 2$, (MP) is simply called a Poincaré condition and the smallest $C_{\mathrm{MP}}$ for which the inequality holds is the Poincaré constant of $\pi$, denoted $C_{\mathrm{P}}$.
\end{definition}


\begin{assumption}\label{assum:lmp}
The target distribution $p(\vx_t)$ satisfy the mirror Poincaré condition with $\phi = \log p(\cdot) + \lambda\norm{\cdot}^2/2$ with constant $C_{\mathrm{LMP}}$, we name this Levenberg-Marquardt Poincaré condition.
\end{assumption}

\begin{proof}
The law $\left(\mu_t\right)_{t \geq 0}$ of LML in Eq. \ref{lml_continuous} satisfies the following Fokker-Planck equation in the weak sense \citep[$\S 5.7$]{karatzas2014brownian}
\begin{equation} \label{}
\begin{aligned}
    \partial_t \mu_t=\operatorname{div}\left(\mu_t\left(\nabla^2 \log p(\cdot)+\lambda \mI\right)^{-1} \nabla \ln \frac{\mu_t}{p}\right)
\end{aligned}
\end{equation} 
which is well-posed with enough regularity \citep[Proposition 6]{bris2008existence}.
Using this, we can compute the derivative of the chi-squared divergence:
\begin{equation} \label{}
\begin{aligned}
& \partial_t \chi^2\left(\mu_t \| p\right) \\
&=\partial_t \int \frac{\mu_t^2}{p}\\
&=2 \int \frac{\mu_t}{p} \partial_t \mu_t\\
&=2 \int \frac{\mu_t}{p} \operatorname{div}\left(\mu_t\left(\nabla^2 \log p(\cdot)+\lambda \mI\right)^{-1} \nabla \ln \frac{\mu_t}{p}\right) \\
& =-2 \int\left\langle\nabla \frac{\mu_t}{p},\left(\nabla^2 \log p(\cdot)+\lambda \mI\right)^{-1} \nabla \ln \frac{\mu_t}{p}\right\rangle \mu_t\\
&=-2 \int\left\langle\nabla \frac{\mu_t}{p},\left(\nabla^2 \log p(\cdot)+\lambda \mI\right)^{-1} \nabla \frac{\mu_t}{p}\right\rangle p
\end{aligned}
\end{equation} 

The Assumption \ref{assum:lmp} implies 
\begin{equation} \label{gron}
\begin{aligned}
\partial_t \chi^2\left(\mu_t \| p\right) & =-2 \int\left\langle\nabla \frac{\mu_t}{p},\left(\nabla^2 \log p(\cdot)+\lambda \mI\right)^{-1} \nabla \frac{\mu_t}{p}\right\rangle p \\
&\geq -2 \frac{1}{C_{\mathrm{LMP}}} \chi^2\left(\mu_t \| p\right)
\end{aligned}
\end{equation} 
Applying Grönwall's inequality to Eq. \ref{gron}, we can get the exponential ergodic convergence rate:
\begin{equation} \label{}
\begin{aligned}
\chi^2\left(\mu_t \| p\right)\leq e^{-\frac{2t}{C_{\mathrm{LMP}}}}\chi^2\left(\mu_0 \| p\right)
\end{aligned}
\end{equation} 

\end{proof}

\begin{figure*}[h]
\centering
\begin{minipage}{0.99\columnwidth}
\centering
\includegraphics[width=\linewidth]{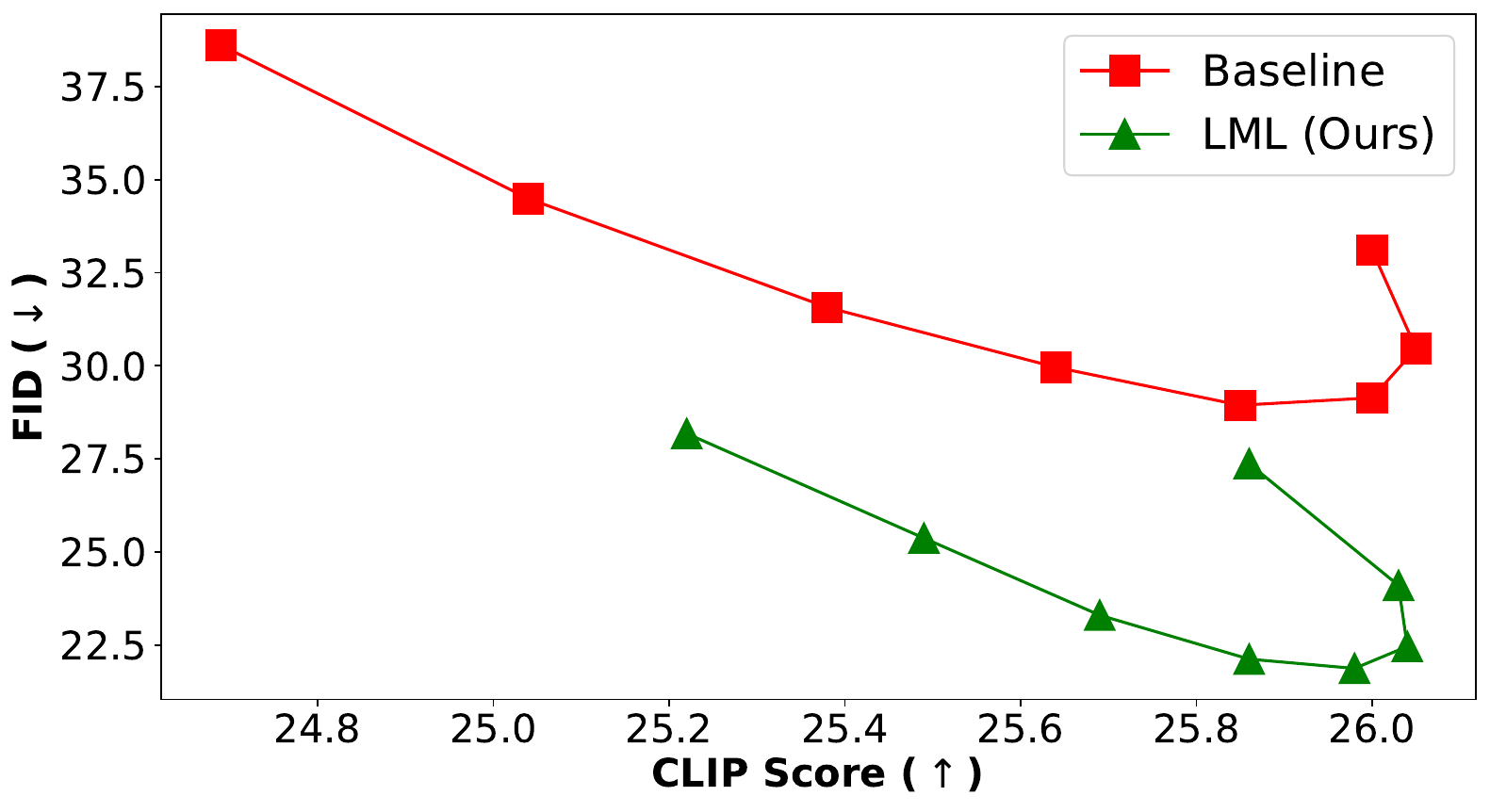}
\subcaption{SD-v1.5}
\end{minipage}
\hfill
\begin{minipage}{0.99\columnwidth}
\centering
\includegraphics[width=\linewidth]{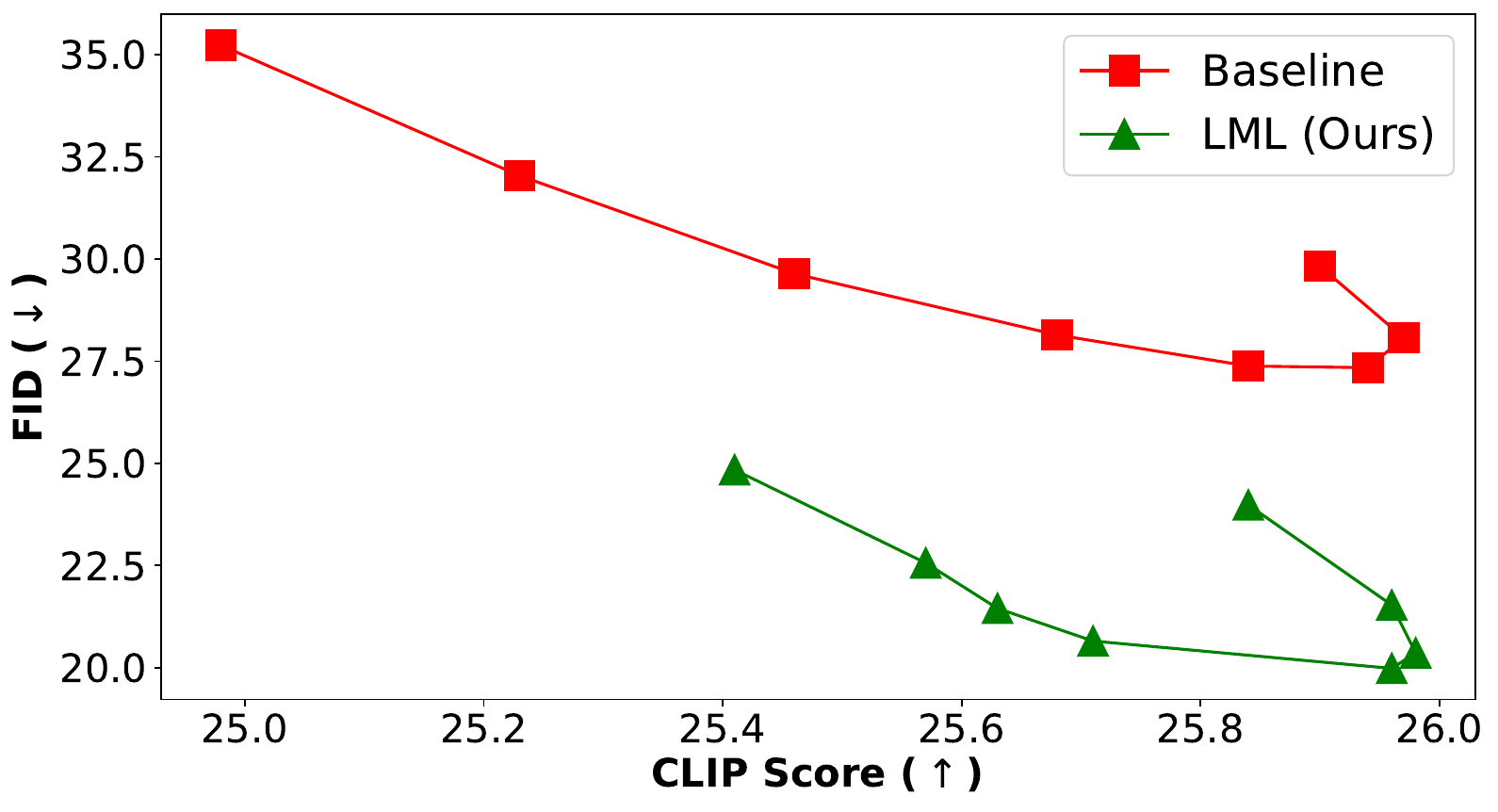}
\subcaption{SD-v2-base}
\end{minipage}
\caption{Comparison of Pareto curves between LML and baseline on SD-v1.5 and SD-v2-base on 30k COCO images, across various guidance scales in [5.0, 6.0, 7.0, 8.0, 9.0, 10.0, 11.0, 12.0], using 5 NFEs.}
\label{fig:pareto}
\end{figure*}

\begin{table*}[t]
\centering
\small
\renewcommand{\arraystretch}{1.1}
\caption{The $\lambda$ setting in Table \ref{tab:unconditional_fid}.}
\resizebox{0.99\textwidth}{!}{
\begin{tabular}{c|cccccccccccc}
\Xhline{1pt}
{\multirow{2}{*}{Methods}} & \multicolumn{9}{c}{$\lambda$ settings on CIFAR-10 generation} \\
\Xcline{2-13}{0.5pt}
& 5 NFEs & 6 NFEs & 7 NFEs & 8 NFEs & 9 NFEs & 10 NFEs& 12 NFEs& 15 NFEs & 20 NFEs & 30 NFEs& 50 NFEs & 100 NFEs \\
\Xhline{1pt}
LML  &  0.0008 & 0.0008 & 0.001 & 0.001 & 0.001 & 0.0008 &0.001&0.001& 0.0005 &0.0003& 0.0001 & 0.00005 \\
\Xhline{1pt}
\end{tabular}
\label{tab:lambda_setting}
}
\end{table*}

\begin{table}[t]
    \centering
    \small
    \renewcommand{\arraystretch}{1.1}
    \caption{The $\lambda$ setting in Table \ref{tab:fid_coco}.}
    \resizebox{0.99\columnwidth}{!}{
    \begin{tabular}{c|cccccccc}
        \Xhline{1pt}
        {\multirow{2}{*}{Methods \textbackslash NFEs}} & \multicolumn{8}{c}{$\lambda$ settings on SD models} \\
        \Xcline{2 - 9}{0.5pt}
        & 5 & 6 & 7 & 8 & 9 & 10 & 12 & 15\\
        \Xhline{1pt}
        \Xcline{1 - 9}{0.5pt}
        &\multicolumn{8}{c}{SD-1.5}\\
        \Xhline{1pt}
        LML & 0.001 & 0.001 & 0.001 & 0.001 & 0.001 & 0.001 & 0.001 & 0.001\\
        \Xhline{1pt}
        \Xcline{1 - 9}{0.5pt}
        &\multicolumn{8}{c}{SD2-base}\\
        \Xhline{1pt}
        LML & 0.001 & 0.001 & 0.001 & 0.001 & 0.001 & 0.001 & 0.001 & 0.0008\\
        \Xhline{1pt}
    \end{tabular}}
    \label{tab:fid_coco_lambda}
\end{table}

\section{Experimental Details}\label{supp:exp_detail}
\subsection{Detail settings}
Across all experiments, for our LML, DPM-Solver, and DPM-Solver++, we maintain the solver order at 3, and the log-SNR trajectory follows the original setup. 
The text prompts of Figure \ref{fig:sd_qualitative_sd15} and \ref{fig:sd_qualitative_sd2b} are taken from \url{https://medium.com/phygital/top-40-useful-prompts-for-stable-diffusion-xl-008c03dd0557}.

\subsection{Hyperparameters}\label{supp:hyperparameter}
\begin{figure}[t]
\centering
\includegraphics[width=0.99\columnwidth]{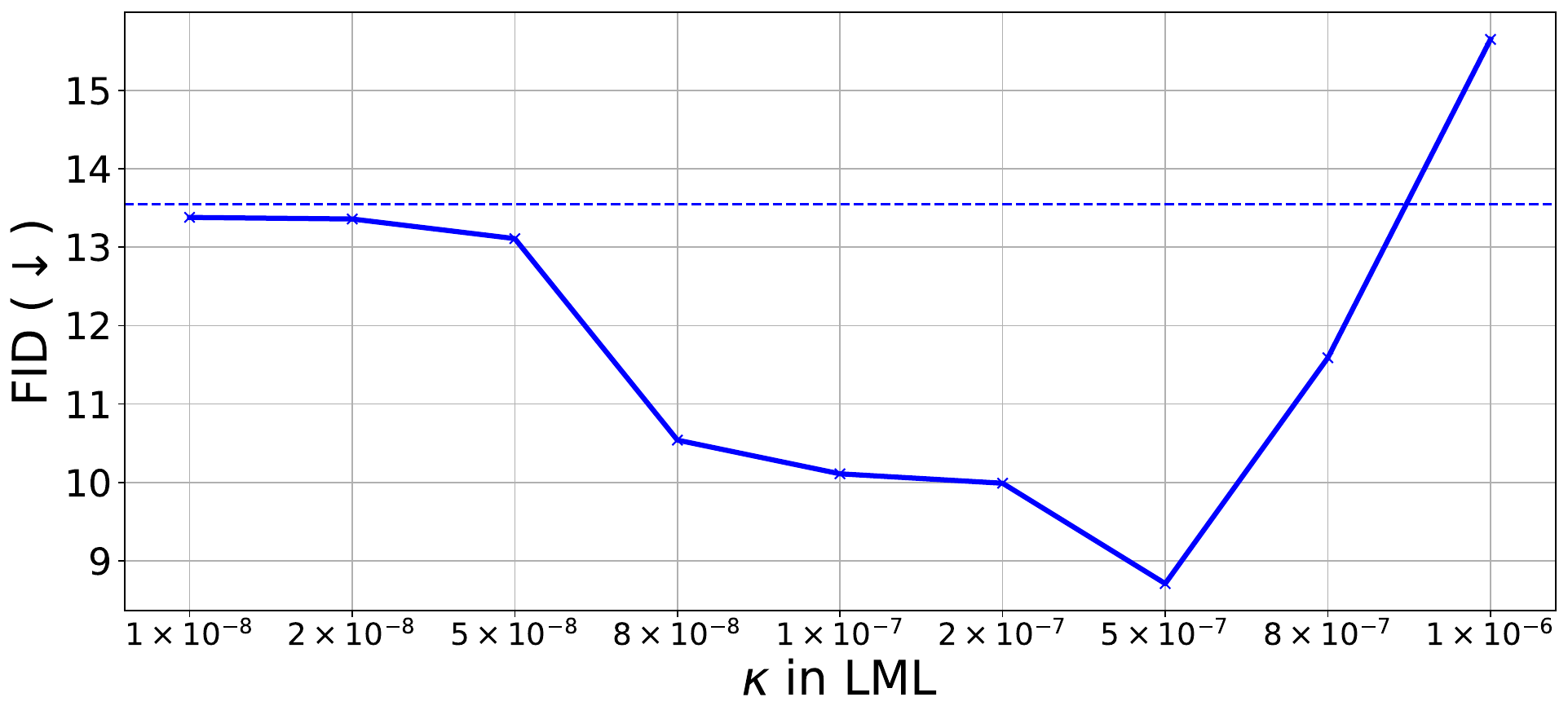} 
\caption{The performance of the LML method on CIFAR-10 generation with various choices of the damping coefficients $\kappa$. The dashed lines signify the performance when $\kappa = 0$. We fix using 10 NFEs and $\lambda = 0.003$. For simplicity, we adopt the DDIM denoise scheme here. }
\label{fig:cifar_kappa_ablation}
\end{figure}

\paragraph{Ablations and settings on $\kappa$}
As illustrated in Figure \ref{fig:cifar_kappa_ablation}, there is a noticeable improvement in performance as $\kappa$ gradually increases. However, excessively large values of $\kappa$ may be detrimental. Due to computational constraints, it is not feasible to optimize $\kappa$ for all of our experiments. Consequently, we have chosen to fix $\kappa = 1\times10^{-8}$ for all tests in Tables \ref{tab:unconditional_fid} and \ref{tab:fid_coco}. This is a very small value that contributes minimally to performance enhancement. There remains considerable potential for performance improvement by fine-tuning $\kappa$ further in our LML sampler.
For CelebA-HQ, we set $\lambda$ as 0.004.
And we set $\lambda$ as 0.001 for SD-XL, 0.006 for PixArt-$\alpha$.

\begin{figure}[t]
\centering
\includegraphics[width=0.99\columnwidth]{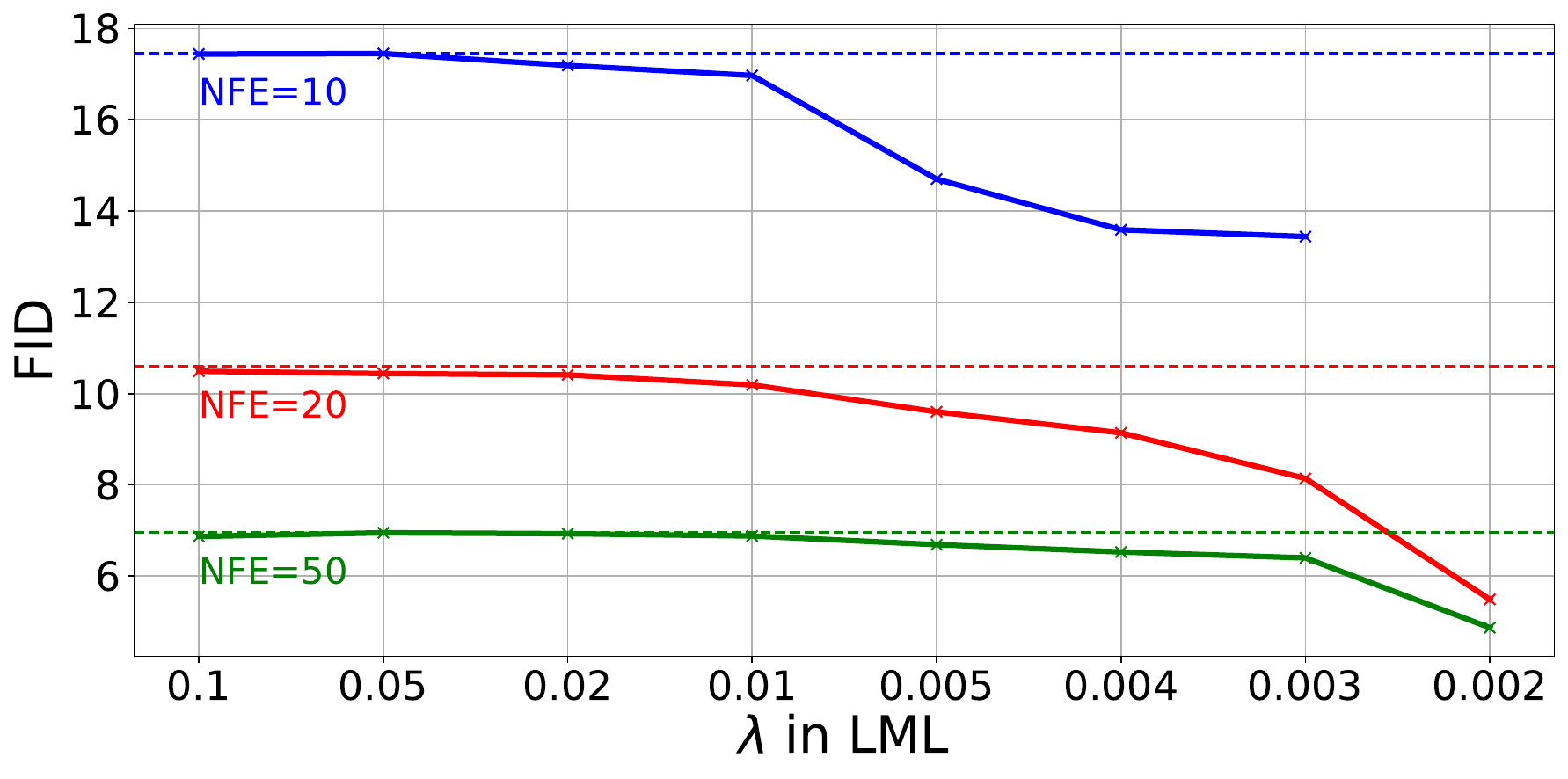} 
\caption{The performance of the LML method on CIFAR-10 generation with various choices of the damping coefficients $\lambda$. The dashed lines signify the performance of the DDIM method. For simplicity, we adopt DDIM denoise scheme here.}
\vspace{-3mm}
\label{fig:cifar_gamma_ablation}
\end{figure}

\paragraph{Settings on $\lambda$}
Due to the behavior of $\lambda$ as in Figure \ref{fig:cifar_gamma_ablation}, we employed a binary search strategy to determine the value for the hyperparameter $\lambda$. In each iteration, we computed the performance of our model using the current $\lambda$ value and then updated the range based on the results. If the performance improved, we would continue the search in the direction of the current $\lambda$; otherwise, we would search in the opposite direction. This process was repeated until we reached 5 iterations. Thus, the total tuning computation budget is controlled. We present the detailed $\lambda$ setting of our experiments in Table \ref{tab:lambda_setting} and \ref{tab:fid_coco_lambda}.

\subsection{Pretrained models}\label{supp:pretrain}
All of the pretrained models used in our research are open-sourced and available online as follows:
\begin{itemize}
    \item ddpm-ema-cifar10
     
     \url{https://github.com/VainF/Diff-Pruning/releases/download/v0.0.1/ddpm_ema_cifar10.zip}

    \item CompVis/ldm-celebahq-256
    
     \url{https://huggingface.co/CompVis/ldm-celebahq-256}
    
     \item stable-diffusion-v1.5
    
     \url{https://huggingface.co/runwayml/stable-diffusion-v1-5}

    \item stable-diffusion-v2-base

     \url{https://huggingface.co/stabilityai/stable-diffusion-2-base}

     \item stable-diffusion-XL

     \url{https://huggingface.co/stabilityai/stable-diffusion-xl-base-1.0}

     \item PixArt-$\alpha$

     \url{https://huggingface.co/PixArt-alpha/PixArt-XL-2-512x512}

\end{itemize}

\section{Additional Experimental Results}\label{supp:addition_results}
\begin{figure*}[h!]
\centering
\begin{minipage}{0.33\textwidth}
\centering
\includegraphics[width=\linewidth]{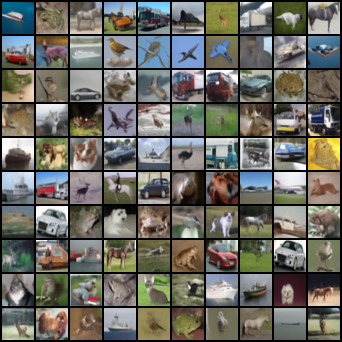}
\subcaption{DDIM \\(FID = 17.45)}
\end{minipage}
\hfill
\begin{minipage}{0.33\textwidth}
\centering
\includegraphics[width=\linewidth]{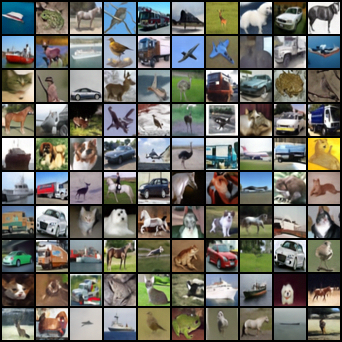}
\subcaption{PNDM \\(FID = 7.16)}
\end{minipage}
\hfill
\begin{minipage}{0.33\textwidth}
\centering
\includegraphics[width=\linewidth]{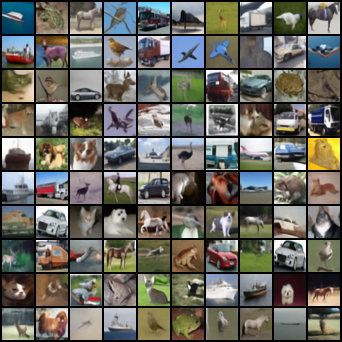}
\subcaption{DPM-Solver \\(FID = 11.13)}
\end{minipage}
\hfill
\begin{minipage}{0.33\textwidth}
\centering
\includegraphics[width=\linewidth]{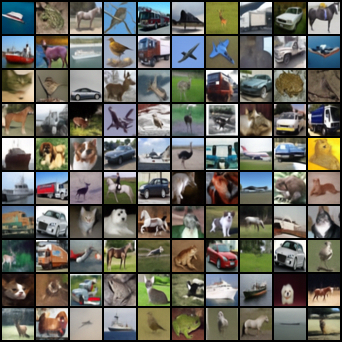}
\subcaption{DPM-Solver++ \\(FID = 10.71)}
\end{minipage}
\hfill
\begin{minipage}{0.33\textwidth}
\centering
\includegraphics[width=\linewidth]{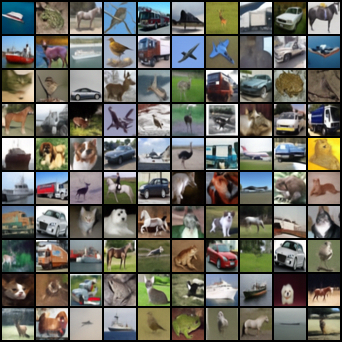}
\subcaption{UniPC \\(FID = 10.70)}
\end{minipage}
\hfill
\begin{minipage}{0.33\textwidth}
\centering
\includegraphics[width=\linewidth]{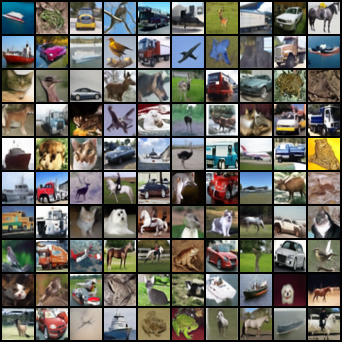}
\subcaption{LML\textbf{(Ours)} \\ (FID = \textbf{6.54})}
\end{minipage}
\caption{Comparison of the CIFAR-10 generation task performance between our LML method and the baseline methods under 10 NFEs. Our LML achieves a superior FID score and enhances visual realism. They are tested using the same pretrained model and seeds.}
\label{fig:cifar_dpm}
\end{figure*}

A visual comparison of LML and the baselines under 10 NFEs with the same seeds is provided in Figure \ref{fig:cifar_dpm}, showing that our LML technique noticeably enhances detail visual realism.
\subsection{CLIP v.s. FID Pareto curve experiment}
Inspired by Imagen \cite{saharia2022photorealistic}, we plot CLIP vs. FID Pareto curves by varying guidance values within the range [5.0, 6.0, 7.0, 8.0, 9.0, 10.0, 11.0, 12.0] in Fig. \ref{fig:pareto}. 
Specifically, Fréchet Inception Distance (FID) \cite{heusel2017gans} calculates the Fréchet distance between the real data and the generated data.
A lower FID implies more realistic generated data.
While the Contrastive Language-Image Pre-training (CLIP) \cite{radford2021learning} score measures the similarity between the generated images and the given prompts. 
A higher CLIP score means the generated images better match the input prompts.
Our LML sampler exhibits substantial improvements over the baseline sampler.
As the guidance scale increases, the LML sampler consistently maintains a lower FID compared to baseline for achieving a similar CLIP score.
This emphasizes that our approach not only enhances image realism but also ensures better adherence to the input prompts.

\subsection{More visual comparisons}
We provide more visual quantitative comparisons of our LML sampler with baseline on CIAFR-10, as shown in Figure \ref{fig:cifar_dpm}.
It is shown that our LML generates samples with enhanced visual fidelity.




\section{Discussions}
\subsection{Social Impacts}
The enhanced image generation sampler proposed in this paper has substantial potential across multiple domains, including machine learning, healthcare, environmental modeling, and economics. However, while this research offers great promise for positive change, it is essential to consider potential adverse societal implications. The improved capabilities of generative models provided by LML might be misused. For example, it could be exploited to generate aesthetically enhanced deepfakes, thereby contributing to the spread of misinformation. In the healthcare sector, if not appropriately regulated, the use of synthetic patient data could give rise to ethical concerns. Thus, it is of utmost importance to ensure that the results of this research are applied ethically and responsibly, with adequate safeguards in place to prevent misuse and protect privacy.

\subsection{Denoising Schemes and Timesteps}
Theoretically, our LML is independent of existing denoising schemes such as DDIM, DPM-Solver, UniPC, and others. Due to restricted computational resources, we currently only integrate DDIM and DPM-Solver with our LML. However, it would be intriguing to incorporate more advanced denoising schemes and timestep selection methods to further improve the performance of our LML.

\subsection{Connections to Mirror Duality} \label{supp:mirror}
In convex optimization, the mirror mechanism \cite{nemirovskij1983problem} is a powerful framework that bridges the primary space (original variable domain) and the dual space (transformed domain via convex conjugation). The primary space hosts the optimization variables, while the dual space, derived through the Legendre-Fenchel transform, captures conjugate representations of convex functions. The Legendre map (or Legendre transform) acts as a bijection between these spaces, converting a convex function in the primary space into its dual counterpart. This duality enables leveraging geometric properties of both spaces to design efficient algorithms.

In optimization, the mirror mechanism is used to either handle a constraint problem or for acceleration. Our method can be viewed as doing a mirror diffusion model \cite{liu2023mirror} in the dual space defined by a Legendre transform map of $\nabla\log p_t(\cdot)+\lambda\norm{\cdot}$ to enhance sampling quality.

\subsection{Limitations}
In this paper, we employ a fixed $\lambda$ for LML throughout all timesteps. 
As demonstrated in advanced LM literature \cite{ngia2000efficient,fan2019adaptive}, a more effective strategy may be to adaptively control $\lambda$.
Additionally, there is potential to develop a more refined rank approximation of the diffusion Hessian by following the concept of the L-BFGS-type method \cite{byrd1995limited}, which could enhance the accuracy of our diffusion Hessian approximation.
Our technique may also enhance the generation quality of flow matching models \cite{lipman2022flow, zhu2024analyzing} or variational inference models \cite{zhu2024neural,wang2024gad}.
Our technique may also enhance the downstream applications of diffusion models in automated driving \cite{tu2025driveditfit}, touch-generation \cite{tu2025texttoucher}, domain-transfer \cite{feng2024lw2g,feng2024pectp}, language generation \cite{fu2025iap}, exact inversion \cite{wang2024belm}, and missing data imputation \cite{chen2024rethinking}.

\end{document}